\documentclass[sigconf]{acmart}
\AtBeginDocument{%
  }

\copyrightyear{2024}
\acmYear{2024}
\setcopyright{acmlicensed}
\acmConference[KDD '24]{Proceedings of the 30th ACM SIGKDD Conference on Knowledge Discovery and Data Mining}{August
25--29, 2024}{Barcelona, Spain}
\acmBooktitle{Proceedings of the 30th ACM SIGKDD Conference on Knowledge
Discovery and Data Mining (KDD '24), August 25--29, 2024, Barcelona, Spain}
\acmDOI{10.1145/3637528.3671776}
\acmISBN{979-8-4007-0490-1/24/08}

\usepackage[normalem]{ulem}
\usepackage{hyperref}
\usepackage{algorithm}
\usepackage{algorithmic}
\usepackage{enumitem}
\usepackage{graphicx} 
\usepackage{float}

\begin{document}

\title{Learning from Emergence: A Study on Proactively Inhibiting the Monosemantic Neurons of Artificial Neural Networks}


\author{Jiachuan WANG}
\orcid{0000-0001-6473-8221}
\affiliation{%
  \institution{HKUST}
  \city{Hong Kong SAR}
  \country{China}}
\email{jwangey@connect.ust.hk}

\author{Shimin DI}
\orcid{0000-0002-7394-0082}
\authornote{The corresponding author.}
\affiliation{%
  \institution{HKUST}
  \city{Hong Kong SAR}
  \country{China}}
\email{sdiaa@connect.ust.hk}

\author{Lei CHEN}
\orcid{0000-0002-8257-5806}
\affiliation{%
  \institution{HKUST(GZ), HKUST}
  \city{Guangzhou}
  \country{China}}
\email{leichen@hkust-gz.edu.cn}

\author{Charles Wang Wai Ng}
\orcid{0000-0001-6693-3151}
\affiliation{%
  \institution{HKUST(GZ), HKUST}
  \city{Guangzhou}
  \country{China}}
\email{charles.ng@ust.hk}


\begin{abstract}

Recently, emergence has received widespread attention from the research community along with the success of large-scale models.
Different from the literature, we hypothesize a key factor that promotes the performance during the increase of scale: the reduction of monosemantic neurons that can only form one-to-one correlations with specific features.
Monosemantic neurons tend to be sparser and have negative impacts on the performance in large models.
Inspired by this insight, we propose an intuitive idea to identify monosemantic neurons and inhibit them.
However, achieving this goal is a non-trivial task as there is no unified quantitative evaluation metric and simply banning monosemantic neurons does not promote polysemanticity in neural networks.
Therefore, we first propose a new metric to measure the monosemanticity of neurons with the guarantee of efficiency for online computation, then introduce a theoretically supported method to suppress monosemantic neurons and proactively promote the ratios of polysemantic neurons in training neural networks.
We validate our conjecture that monosemanticity brings about performance change at different model scales on a variety of neural networks and benchmark datasets
in different areas, including language, image, and physics simulation tasks.
Further experiments validate our analysis and theory regarding the inhibition of monosemanticity.

\end{abstract}



\maketitle

\section{Introduction}
\label{sec:intro}

\begin{figure}[!t]
\centerline{\includegraphics[width=\columnwidth]{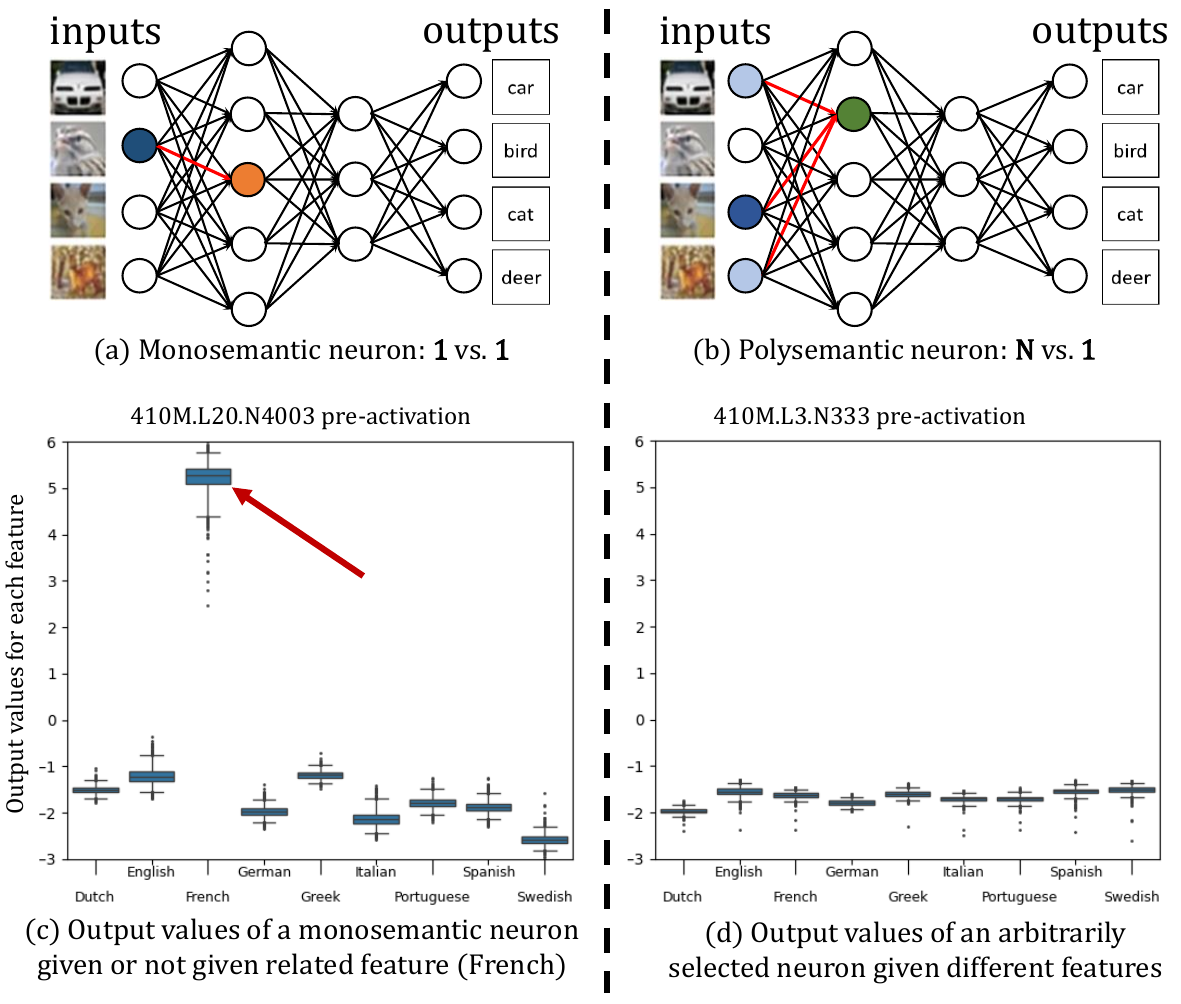}}
\vspace{-5px}
\caption{
Demonstration of important concepts with statistics:
(a) A monosemantic neuron (orange) ideally activates for one specific type of feature.
(b) A polysemantic neuron (green) activates for multiple features.
(c) The output values of a monosemantic neuron when different features are inputted. Its related feature (French) produces values that significantly stand out from other features.
(d) The output values of an arbitrarily selected neuron (layer 3, number 333) given different features. The values fluctuate slightly with similar patterns. These statistics are obtained by inspecting the Pythia-v0 410M model \cite{pythia}.}
\label{fig:motivation}
\vspace{-10px}
\end{figure}

\begin{figure*}[!t]
\begin{center}
\centerline{\includegraphics[width=2\columnwidth]{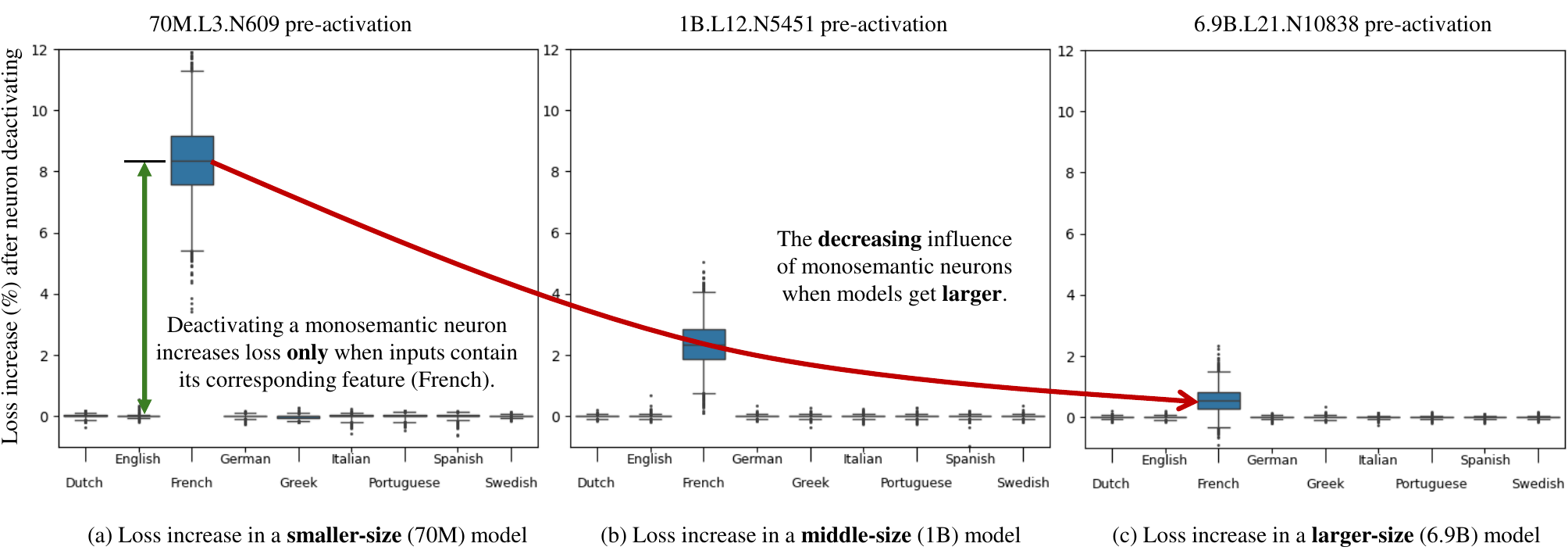}}
\vspace{-5px}
\caption{
We detect the monosemantic neurons of ``French'' following the sparse probing paper~\cite{sparseprobe}
and run the experiments on Pythia-v0~\cite{pythia}.
After deactivating a monosemantic neuron for ``French'', there is an increase in the loss given inputs of different language features (e.g., Dutch and Greek) on Pythia models of different scales:
(a) on the 70M Pythia-v0 model,
(b) on the 1B Pythia-v0 model,
and (c) on the 6.9B Pythia-v0 model.
It can be observed that these neurons are typically monosemantic, causing a large increase in loss only when the input contains ``French'' (see green arrow). However, for larger models, deactivation of these neurons leads to a smaller increase in loss (see red arrow). This gives us a hint that monosemanticity may be negatively related to the scale and performance of larger models.
}
\vspace{-15px}
\label{fig:loss}
\end{center}
\end{figure*}

The activity of artificial neural networks diminished for decades before experiencing great success after the 2010s \cite{whatneuron, alex}.  
One major difference compared to previous models is the increasing scale.
In recent years, large neural networks have much larger scales in terms of datasets, model sizes, and training quantities, which have achieved remarkable results in various fields \cite{GPT3,LLAMA2}. 
Given increasing scales, 
the performance improvements can usually be estimated under the scaling law~\cite{scaling}, which follows stable but slow increasing trend of power laws~\cite{GPT4}. 
However, improvements that significantly deviate from estimation occur when the scale increases.
Emergence, an intriguing phenomenon in large-scale models, refers to the gradual improvement of model performance before the scale reaching a certain threshold, followed by a rapid enhancement once the threshold is surpassed~\cite{DBLP:journals/tmlr/WeiTBRZBYBZMCHVLDF22}. 
Increasing evidence suggests that the surprises may not arise from new module and architecture designs, but rather from the underlying nature of scale changes~\cite{DBLP:journals/corr/abs-2311-00237}. 
A critical question arises: 
\textbf{\textit{people increase the model scale and get better results, but what has changed underlying the process?}}

From the perspectives of \textit{monosemantic} and \textit{polysemantic} neurons,
some pioneer works try to interpret the performance from small to large-scale models. 
Through statistical analysis of the relationships between neurons and input features, a neuron is considered monosemantic if it forms a 1-1 correlation with its related input feature~\cite{DBLP:journals/pnas/BauZSLZ020, softmax} (Figure~\ref{fig:motivation}(a)). 
In contrast, polysemantic neurons activate for several features that are weakly correlated \cite{bricken2023towards, circuits, goh2021multimodal, softmax} (Figure~\ref{fig:motivation}(b)).
By comparing Figure~\ref{fig:motivation}(c) and (d), we can observe that the activation level of a monosemantic neuron for the feature ``French'' is significantly different from that of the neuron for other features and from that of other non-monosemantic neurons for ``French''.
Researchers find that smaller models incur larger errors~\cite{sparseprobe} when an input's related monosemantic neuron is disabled.
Following the favor of monosemantic neurons from studies on small-scale models, 
existing works focus on enhancing or extracting monosemanticity~\cite{sparseprobe,softmax,bricken2023towards}.
However, 
some evidences support that monosemantic neurons are sparser in larger models~\cite{sparseprobe, bricken2023towards}.
Those observations imply a positive relationship between the decrease in monosemanticity and the increase in model scale.
In other words, monosemanticity could be negatively related to large-scale and good performance.
In Figure~\ref{fig:loss}, we conduct experiments to show that, when a monosemantic neuron for ``French'' is deactivated, a smaller model has a greater increase in loss than a larger model.
In this paper,
inspired by those observations, we propose an important hypothesis that 
\textbf{the decrease of monosemantic neurons is a key factor in achieving higher performance as the model scale increases}.
Figure~\ref{fig:analogy} in Appendix~\ref{appx:example2} 
shows an example analogy to demonstrate our idea.

Since the research community does not realize the above hypothesis, 
we rather conclude the current paradigm of training neural networks as a \textbf{passive} process in decreasing monosemantic neurons.
This raises an interesting question: 
\textit{can we \textbf{proactively} induce the decrease of monosemantic neurons in artificial neural networks to achieve high performance?}

In this paper, we propose to learn from emergence and present a study on proactively inhibiting the monosemantic neurons, which is achieved in two phases:
(i) monosemantic neuron detection and (ii) monosemantic neuron inhibition.
Unfortunately, it is challenging to design detection and inhibition methods.
First, strictly defining monosemantic neurons is still under discussion in quantitative analysis, i.e., detection is impossible without a consensual definition.
Besides, existing works~\cite{sparseprobe, bricken2023towards, bills2023language} for detecting the activation degree of neurons introduce extra computation and may suffer from the efficiency issue.
Second, we prove that simply prohibiting the activation of monosemantic neurons will intensify the monosemanticity of artificial neural networks.
To solve these challenges, we first propose a new metric to measure the monosemanticity of neurons with the guarantee of efficiency for online computation. 
Then, we identify the drawbacks of simple methods and propose a theoretically supported Reverse Deactivation method to suppress monosemantic neurons and promote the ratios of polysemantic neurons in training neural networks.
Our contributions are summarized as follows:
\begin{itemize}[leftmargin=*]
\item Inspired by emergence, we propose a novel idea to study the impact of monosemantic and polysemantic neurons on the performance of artificial neural networks.
Different from the literature, we propose a hypothesis that monosemanticity could be negatively related to good performance on large-scale models.

\item There is no quantitative definition of monosemanticity with high computational efficiency. 
To effectively detect monosemantic neurons, we propose a new evaluation metric of monosemanticity with a theoretical guarantee on online computation.

\item To overcome the inherent drawback of the naive inhibition methods,
we propose a theoretically supported reverse deactivation method to suppress monosemantic neurons.
It can be integrated as a parameter-free and flexible insertion module, which can adapt to various neural network structures and introduce no computational overhead during testing.

\item To validate the effectiveness and generalizability of our method, we conducted tests on various tasks including language, image, and physics simulation. 
The proposed method MEmeL is applied to milestone models such as Bert, ConvRNN, and different architectures like Transformer, CNN, and RNN. The experimental results demonstrate that MEmeL performs well on different neural networks and corresponding tasks.
\end{itemize}


\section{Preliminary}
\label{sec:pre}

Given a training set $D=\{(\mathbf{x},\mathbf{y})\}$ of labeled examples, artificial neural networks aim to train a neural network model $f(\mathbf{W};\mathbf{x})$ parameterized by $\mathbf{W}$ so that the predicted output of $\mathbf{x}$ is close to the ground truth $\mathbf{y}$. 
Under the supervised learning task, we take the cross entropy function $\mathcal{L}(\mathbf{W};D)$ as the example:
\begin{equation}
    \min_{\mathbf{W}} \mathcal{L}(\mathbf{W};D) = \min_{\mathbf{W}} - \sum_{(\mathbf{x},\mathbf{y})\in D}\mathbf{y}\log f(\mathbf{W};\mathbf{x}),
\end{equation}
where $\mathbf{x}=[x_1,\cdots,x_d]\in\mathbb{R}^d$ is a $d$-dimensional vector representing the $n$-th input data sample, such as a weather record series covering $d$ days or an image with height $\times$ width $=d$ pixels.
Without loss of generality, let $f(\mathbf{W};\mathbf{x})$ be a $L$-th layer fully connected neural networks~\cite{nnfoundation}.
We denote $h_{i}^{\ell}$, $\sigma_{i}^{\ell}(\cdot)$, $z_{i}^{\ell}$ as the input after linear transformation, activation function, and output of the $i$-th hidden unit at $\ell$-th layer, respectively.
Then, we may formally define the computation within neurons as follows:
\begin{align}
& h_j^{\ell} = \sum_{i} w_{ij}^{\ell}z_{i}^{\ell-1},\\
     & z_{i}^{\ell} = \sigma_{i}^{\ell}(h_{i}^{\ell}), 
\end{align}
where 
$z_i^0\in\mathbf{z}^0=\mathbf{x}$,
$w_{ij}^{\ell}$ denotes the weight of $i$-th hidden neuron at $(\ell-1)$-th layer to $j$-th hidden neuron at $\ell$-th layer. 
We use $\mathbf{o}=\mathbf{z}^L$ to denote the model output.
Existing works on monosemanticity mainly study the layers of values outputted by activation functions (i.e., $\mathbf{z}^\ell$). In contrast to other linear layers, these activated values undergo element-wise nonlinearity and are more likely to have independent meanings \cite{sparseprobe, embeddingkv}. 
Thus, 
without loss of generality, we zoom in and study monosemanticity based on one single layer of activated values, e.g., $\mathbf{z}^\ell$ in the $\ell$-th layer.
Then,  $f(\mathbf{W};\mathbf{x})$ can be divided into a frontal model $f_1(\cdot)$ and a followed model $f_2(\cdot)$ at $\ell$-th layer of output as follows:
\begin{align}
&\mathbf{z} = f_1(\mathbf{x}) =  \sigma^\ell(\mathbf{W}^\ell(\cdots\sigma^2(\mathbf{W}^2\sigma^1(\mathbf{W}^1\mathbf{x}))\cdots)),
\\
&\mathbf{o} = f_2(\mathbf{z},\mathbf{x})=\text{Softmax}(\mathbf{W}^L\cdots\sigma^{\ell+2}(\mathbf{W}^{\ell+2}\sigma^{\ell+1}(\mathbf{W}^{\ell+1}\mathbf{z}))\cdots),
\label{eq:followed}\end{align}
where the frontal model $f_1(\cdot)$ takes the original data $\mathbf{x}$ as inputs and delivers output values in $\ell$-th layer $\mathbf{z}$ (with the superscript $\ell$ omitted for simplicity in notation.) 
As the rest components of the network,
the following model $f_2(\cdot)$ takes $\mathbf{x}$ and $\mathbf{z}$ and outputs $\mathbf{o}$ (the entire model's output),
and $\mathbf{W}^{\ell}\in \mathbb{R}^{d_{\ell}\times d_{\ell+1}}$ 
denotes weights of linear transformation from the $\ell$-th layer to the $(\ell+1)$-th layer.

\subsection{Activation and Monosemantic}

Despite the progress of neural networks and emergence, there is no consensus definition for an ``activated'' neuron and a ``monosemantic'' neuron \cite{softmax}.
Here we try to summarize an intuitive introduction to describe them.

\noindent
\textbf{The Concept of Activation:}
If an input $\mathbf{x}^{[n]}$ triggers a neuron $z_i$ to output a value $f_1(\mathbf{x}^{[n]})_i$ that deviates ``significantly'' from the statistical mean (i.e., $\Bar{z}_i$)
, we say that neuron $z_i$ is activated by input $\mathbf{x}^{[n]}$. The challenge in defining activation lies in reaching a consensus on the meaning of ``significantly". Instead, we can provide a relative definition as follows.
Generally, $i$-th neuron at $\ell$-th layer is considered to be more activated on $\mathbf{x}^{[2]}$ if:
\begin{equation}
\label{eq:relative}
\left\Vert \Bar{z}_i-(f_1(\mathbf{x}^{[1]}))_i\right\Vert<\left\Vert\Bar{z}_i-(f_1(\mathbf{x}^{[2]}))_i\right\Vert,
\end{equation}
where $\mathbf{x}^{[{n}]}\in D$ is the ${n}$-th sample in the training set $D$,
$\Bar{z}_i$ is the mean value of $i$-th neuron given all training samples $D$ at $\ell$-th layer,
$(f_1(\mathbf{x}^{[1]}))_i$
denotes the $i$-th output of the frontal model (i.e., $z^{\ell}_i$ for $\mathbf{x}^{[1]}$),
and $\left\Vert\cdot\right\Vert$ is a distance metric.
However, while this relative definition is accurate for illustration purposes, it is not concise and is difficult to use for further analysis.

Thus, to give a one-input-wise definition, one may rely on a threshold to define whether a neuron is activated, such as a hard threshold $\tau$.
If the deviation of activation value from the mean $\Vert{\Bar{z}_i-(f_1(\mathbf{x}^{[n]}))_i}\Vert$ exceeds $\tau$, it is generally considered that the neuron is activated. Otherwise, it is in an inactive state.
The reason for setting $\tau$ is that neurons will generally have certain fluctuating outputs even for those unrelated different inputs.
Unfortunately, setting a universal or adaptive $\tau$ remains to be explored. 
Therefore, it is also impractical to quantify activation by the threshold. 

\noindent
\textbf{The Concept of Monosemanticity:}
However, further illustration for monosemanticity is closely related to a definition of activation that considers only one input at a time. We provide an abstract definition for it: $act(z_i, \mathbf{x}^{[n]})$, which equals 1 when $z_i$ is activated by $\mathbf{x}^{[n]}$, and 0 otherwise. One can use a task-oriented definition as implementation.

To understand neural networks, an important research direction is to correlate neuron activations with human-interpretable features, such as Python and German for language processing; fur and grass for image processing, et al.. Existing works construct feature datasets $\{C_1, \cdots, C_m\}$ for $m$ features, each containing a set of inputs. These feature datasets are specifically designed to partition the inputs $X=\{\mathbf{x}\}$, mathematically: 
$$\forall_{i\neq j} C_i\cap C_j=\emptyset; \bigcup_{i=1}^m C_i=X.$$
A neuron $z_i$ is ``monosemantic" if it is only activated on inputs that share a specific feature $C_j$, {that is:
$$\forall_\mathbf{x} act(z_i, \mathbf{x})=1, \mathbf{x}\in C_j; \forall_\mathbf{x} act(z_i, \mathbf{x})=0, \mathbf{x}\notin C_j$$

}
However, features are human-defined and vary a lot. In the previous study, \citet{sparseprobe} considered about 100 features, while \citet{bricken2023towards} studied up to $10^5$ features to fully capture monosemanticity. Without unified feature datasets, it is also difficult to explicitly formalize the definition of ``monosemantic".
More related works are discussed in Appendix~\ref{sec:related}.

Thus, to detect monosemantic neurons, previous studies require manually labeled feature datasets and time-consuming offline computations after model training. To detect and inhibit monosemantic neurons during training, it is necessary to define a lightweight online metric $\phi(\cdot)$ that 
{does not rely on feature datasets,}
which indicates the level of monosemanticity of a neuron.



\subsection{Monosemanticity Inhibition}

To achieve the desired output $z$, deep learning models use optimization strategy $\omega$ to update parameters ($\mathbf{W}$), such as minimizing the explicit loss function through gradient descent. 

However, to inhibit monosemanticity, there is no related loss function to minimize. In this paper, we achieve this objective in two phases.

Recall that the model $f$ is parameterized by $\mathbf{W}$ and split into a frontal model $f_1$ and a followed model $f_2$ with respect to the studied layer of neurons $\mathbf{z}$.  

Assume that we feed input $\mathbf{x}$ to $f$ and find $z_i\in \mathbf{z}$ is monosemantic. By using a well-designed optimization strategy $\omega$, parameters are updated to $\mathbf{W}^*$.
By feeding the same input $\mathbf{x}$ into the neural network, the frontal model, the followed model, and the layer of neurons are updated to $f_1^*$, $f_2^*$, and $\mathbf{z}^*$, respectively. We hope that:
\begin{itemize}[leftmargin=*]
    \item With updated $f_1$, neuron $z_i$ is less activated for input $\mathbf{x}$. Formally, given old $z_i\in \mathbf{z}=f_1(\mathbf{x})$ and updated $z_i^*\in \mathbf{z}^*=f_1^*(\mathbf{x})$, we expect:
    $$\phi(z_i^*)<\phi(z_i).$$
    \item With updated $f_2$, the output is still robust when neuron $z_i$ is deactivated.
    Formally, replace $z_i$ with a weakly
activated value $z_i'$ (i.e., $\phi(z_i') < \phi(z_i)$) to obtain $\mathbf{z}'$, we expect the loss $\mathcal{L}$ satisfies:
$$\mathcal{L}(f_2^*(\mathbf{z}',\mathbf{x})) < \mathcal{L}(f_2(\mathbf{z}',\mathbf{x})).$$      
\end{itemize}

The two phases (1) prevent the \textbf{neuron} from exclusively serving only \textbf{one feature} type, and (2) prevent this \textbf{feature} modeling from relying solely on \textbf{one neuron}.

\section{Methods}
\label{sec:method}

As discussed in Sec.~\ref{sec:intro}, the existing neural network training paradigm is a passive process of decreasing monosemantic neurons.
To proactively reduce monosemanticity, it is an intuitive solution to detect monosemantic neurons and inhibit them.
Unfortunately, achieving these goals is challenging because the monosemanticity measurement does not exist and the simple inhibition method is counterproductive.
Therefore, we first propose a new metric to measure the monosemanticity of artificial neurons with the guarantee of computation efficiency,
then introduce a theoretically supported inhibition method to suppress monosemantic neurons to proactively reduce the ratios of monosemantic neurons in training artificial neural networks.

\subsection{Metric for Monosemanticity}
\label{subsec:method:detect}

To proactively detect monosemantic neurons, it is important to design a metric that is general for different tasks and efficient to calculate.
However, as discussed in Sec.~\ref{sec:pre}, strictly defining ``activated'' is still under discussion in quantitative analysis~\cite{softmax}, which in turn leads to the difficulty of defining ``monosemantic''.
Although some pioneer works explore the measurement of monosemanticity, these metrics are inflexible for different tasks since they require a predefined and manually labeled feature dataset~\cite{sparseprobe,bricken2023towards}.
To address these limitations, 
we propose a robust metric $\phi(\cdot)$ that can accurately detect monosemantic neurons in this subsection. 
This metric fulfills two important criteria: (1) generality to make the metric do not rely on any specific dataset,
(2) efficiency to enable fast online detection during training.

Intuitively, defining monosemantic neurons mainly requires starting from two principles:
high deviation of activation value and low frequency of activation.
First, a neuron is considered ``activated'' when its output value for the current input deviates from the mean value of outputs for all possible inputs. 
{For a monosemantic neuron, its value distribution is more skewed and incurs large deviation \cite{softmax}.}
Second, each monosemantic neuron only activates when its corresponding feature is inputted,
which rarely happens considering the current datasets with steadily growing scales of samples and feature types.
Following two principles, we formally define our metric Monosemantic Scale (MS for short) as follows.
\begin{definition}[Monosemantic Scale]
Given a neuron $z_i\in \mathbf{z}$, we denote its historical samples under $m$ inputs $\{\mathbf{x}^{[1]},\mathbf{x}^{[2]},\cdots,\mathbf{x}^{[m]}\}$ as $\{z^{[1]}_i,z^{[2]}_i,\cdots,z^{[m]}_i\}$ and new value under the $(m+1)$-th input $\mathbf{x}^{[m+1]}$ as $z^{[m+1]}_i$. The Monosemantic Scale is defined as:
\begin{equation}
\label{eq:phi}
    \phi(z^{[m+1]}_i) =\frac{(z^{[m+1]}_i-\Bar{z}_i)^2}{S^2},
\end{equation}
where $$\Bar{z}_i=\frac{\sum_{j=1}^mz^{[j]}_i}{m}, S^2=\frac{\sum_{j=1}^m(z^{[j]}_i-\Bar{z}_i)^2}{m-1},$$
are the sample mean and sample variance, respectively. 
\end{definition}
In the following contents, we will focus on this single neuron and use $z$ for simplicity.

As shown in Eq.~\eqref{eq:phi}, the measurement $\phi(\cdot)$ is proportional to the degree of deviation from the mean. The high deviation of the activation value ensures that 
the term $(z^{[m+1]}-\Bar{z})^2$ in $\phi$ can effectively identify neurons with high monosemanticity. 
Besides, the metric is inversely proportional to the degree of fluctuation because the size of the deviation is also highly correlated with the fluctuation of the activation value on the current data set.
Since $\Bar{z}_i$ is mainly determined by the values when the neuron is deactivated,
using the mean as the benchmark for evaluating deviations ensures that the defined neurons with high activation values are rare, i.e., low frequency.

As discussed in Sec.~\ref{sec:intro} and \ref{sec:pre}, 
prior works need to first find the neuron-feature relationships 
\cite{sparseprobe,bricken2023towards} under the manually defined feature data set.
Instead, we relax the requirement of discovering corresponding features
and eliminate the need for a predefined feature set since we focus on finding monosemantic neurons.

\begin{figure*}[t!]
\vskip 0.2in
\begin{center}
\centerline{\includegraphics[width=2\columnwidth]{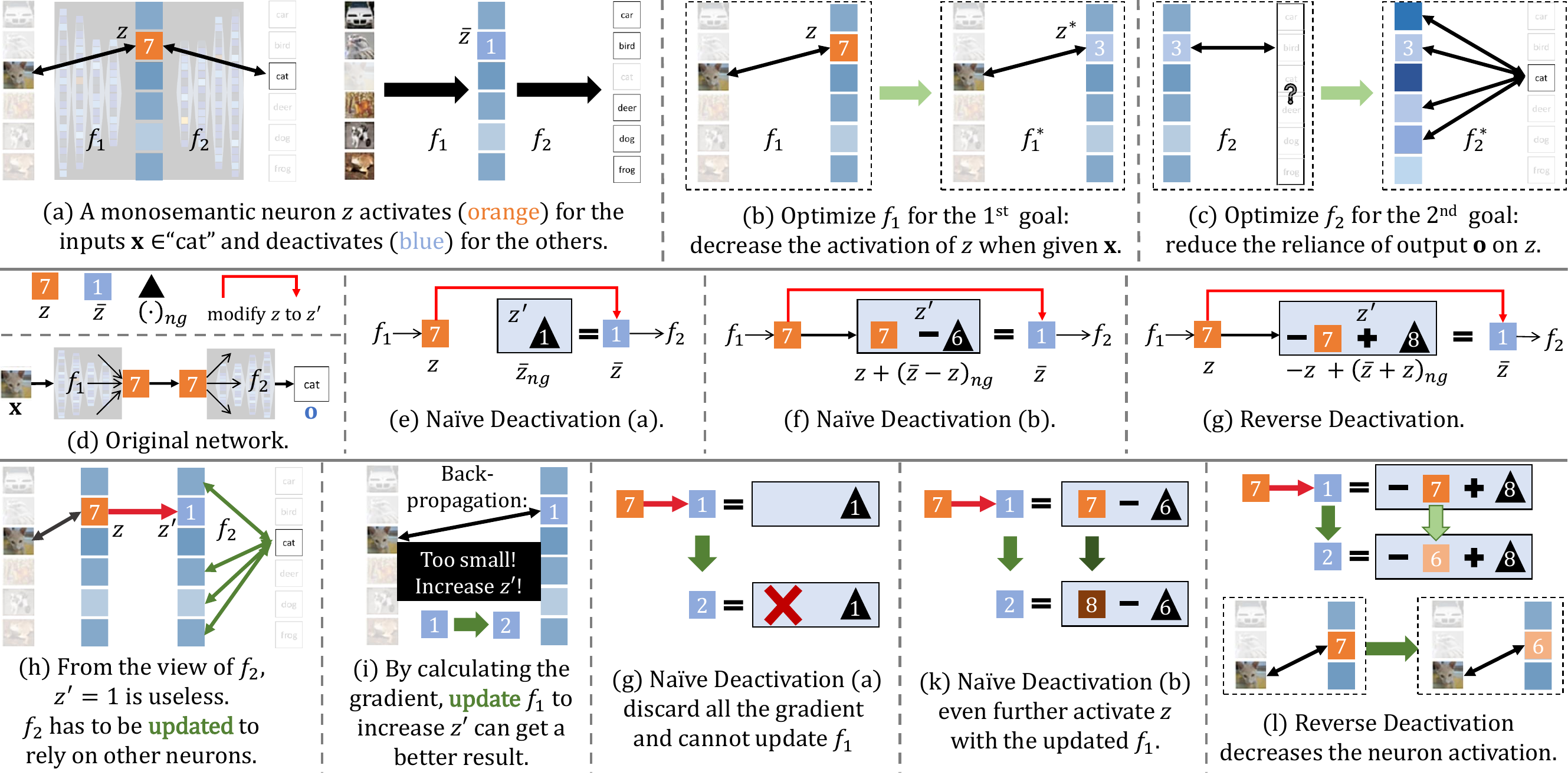}}
\caption{Illustration of problems and solutions to inhibit monosemanticity.
(a) A monosemantic neuron $z$ only activates (orange) for the feature ``cat" with a high mean value ($=7$). $z$ is deactivated (blue) for other inputs with a small mean value ($\Bar{z}=1$).
(b) The first goal is to optimize the frontal model $f_1$ so that $z$ is less activated given the input ``cat".
(c) The second goal is to optimize the followed model $f_2$ so that a correct output for ``cat" does not solely rely on $z$.
(d) Zoom in on the original model at neuron $z$. 
(e) Naive solution that sets $z'$ to a constant 1 without gradient.
(f) Naive solution that decreases the value of $z$ to $\Bar{z}$ with a constant 6 without gradient.
(g) Reverse Deactivation that first reverses $z$ then pushes the output value to $\Bar{z}$ by adding a constant 8 without gradient.
(h) All the methods can achieve the second goal by outputting a value $\mathcal{V}(z')=\Bar{z}$ to $f_2$. As $\Bar{z}$ provides little information, $f_2$ must learn to rely on other neurons.
(i) When calculating the gradient, $f_1$ will find that $z$ is too small and tends to increase it (e.g., from 1 to 2).
(j) Naive method (a) cannot update related parameters without gradient. 
(k) Naive method (b) further increases the underlying $z$ activation (7 to 8).
(l) Reverse Deactivation inherently deactivates $z$ (from 7 to 6). When a new batch arrives, the updated $z^*$ activates less (=6) for ``cat" compared with $z$.}
\label{fig:RD}
\end{center}
\end{figure*}

\noindent
\textbf{Metric Online Computing Guarantee.}
It is mandatory to compute the metric $\phi(\cdot)$ to avoid excessive computational burden caused by detecting monosemanticity.
As discussed in Sec.~\ref{sec:pre}, pioneer works (e.g., pairwise comparison in Eq.~\eqref{eq:relative}) and other detection methods like probes necessitate training and offline inference for statistical analysis may be costly for online training.
Here we show that for each neuron, our proposed MS $\phi(\cdot)$ can be obtained by keeping track of 2 variables in constant time ($O(1)$). Since inputs are typically received in batches (with the number of new samples being greater than 1) during the training of deep learning models, we present the following lemma for training with a batch size of $b$.

\begin{lemma}
\label{lemma:phib}
Denote $\mu_m$ as the value of the sample mean $\Bar{z}$ given $m$ samples, while $\upsilon_m$ as the sample variance $S^2$. When the $(m+1)^{th}\sim(m+b)^{th}$ samples ${z^{[m+1]}, \cdots, z^{[m+b]}}$ come, one can obtain the updated values via:
\begin{align}
    \mu_{m+b}&=\frac{m\mu_m+b\mu_b'}{m+b},\\
    \upsilon_{m+b}&=\frac{mb(\mu_m-\mu_b')^2}{(m+b-1)(m+b)}+\frac{b\upsilon'_b+(m-1)\upsilon_m}{m+b-1},
\end{align}
where $\mu_b'=\frac{\sum_{i=1}^bz_{[m+i]}}{b}$ and $\upsilon'_b=\frac{\sum_{i=1}^b(z_{[m+i]}-\mu'_b)^2}{b}$, 
which is of $O(1)$ time and memory complexity as $b$ is a constant. 
\end{lemma}

The proof is given in Appendix~\ref{appx:phib}.
In the implementation, we introduce a forgetting mechanism during model updates, where the influence of previous samples should decay.
For details, please refer to Algorithm~\ref{alg:metric} in Appendix~\ref{appx:metric}.

\subsection{Inhibition of Monosemanticity}
\label{subsec:method:inhibit}

After defining a quantitative metric in Sec.~\ref{subsec:method:detect},
it is intuitive to inhibit those monosemantic neurons directly.
Unfortunately, simply deactivating monosemantic neurons will intensify the monosemanticity of neural networks.
In this subsection, 
we will first prove the weakness of the native solution in Sec.~\ref{sssec:naive}.
To address this unexpected phenomenon,
we propose a theoretically supported reverse deactivation method in Sec.~\ref{sssec:reverse}.

\subsubsection{Naive Deactivation}
\label{sssec:naive}

Recall that given a monosemantic neuron $z$ and an input $\mathbf{x}$ that contains its exclusive feature, we aim to optimize the model in two aspects: 
(1) decrease the activation level of $z$ when given $\mathbf{x}$ (Figure~\ref{fig:RD}(b)); and (2) reduce the reliance of output $\mathbf{o}$ on the activation of $z$ (Figure~\ref{fig:RD}(c)).

Without loss of generality, 
we focus on the most monosemantic single neuron $z$ in a layer of neurons $\mathbf{z}$.
As defined in Sec.~\ref{subsec:method:detect}, such a neuron has a large 
{$\phi(z)=(z-\Bar{z})^2/S^2$}, 
which is expected to be inhibited during model optimization. 
From the perspective of information theory, suppose that the distribution of activation values follows a normal distribution, a neuron provides the least amount of information $I(z)$ when its value equals its statistical mean:
\[
\min_z\mathbb{E}[{I(z)}]=\min_z\mathbb{E}[{-\log(P(z))}]=-\mathbb{E}\log(P(\Bar{z})),
\]
which is also its most inactive state. $P(\cdot)$ represents the probability density function of values of $z$.

{
Thus, a straightforward idea to deactivate $z$ is to modify its value to $\Bar{z}$. We denote the modified neuron as $z'$.} One can find two naive solutions to deactivate a neuron:
{
\begin{align}
\label{eq:naive}
\text{way}(a):\ \ z'= \ &\Bar{z}_{ng},\\
\text{way}(b):\ \ z'= \ &z+(\Bar{z}-z)_{ng},
\end{align}}
where subscript $\cdot_{ng}$ refers to a ``no gradient" fixed-value scalar without trainable parameters, compared with $z\in \mathbf{z}=f_1(\mathbf{x})$ which is updatable.
{
Two examples are given in Figure~\ref{fig:RD}(e,f).
Both solutions ensure that $f_2$ receives an updated $z'$, in which the value of $\mathcal{V}(z')=\Bar{z}$ provides little information. As $z'$ is valueless, $f_2$ has to adjust its parameter to utilize information from other neurons in $\mathbf{z}$ for good output $\mathbf{o}^*=f_2^*(\mathbf{z}', \mathbf{x})$. Such a process achieves our second goal of reducing the dependence of output $\mathbf{o}$ on the activation of $z$ (Figure~\ref{fig:RD}(l)).
}

\begin{figure*}[t!]
\vskip 0.2in
\begin{center}
\centerline{\includegraphics[width=2\columnwidth]{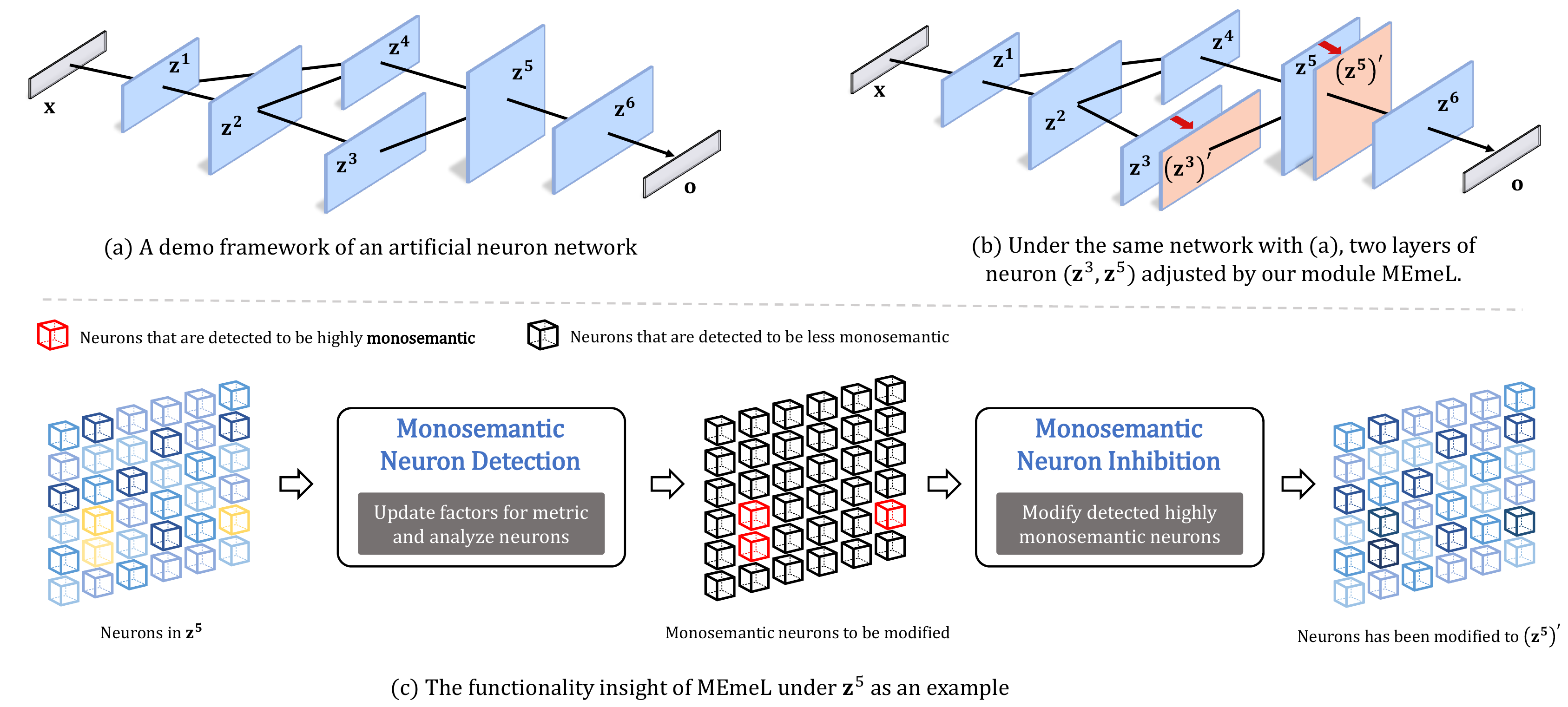}}
\caption{Overview of our method:
(a) An arbitrary neural network framework. $\mathbf{x}$ represents the input and $\mathbf{o}$ represents the output. $\mathbf{z}$s represent hidden layers of neurons, and arrows indicate the dependency relationships.
(b) Our module is inserted after $\mathbf{z}^{3}$ and $\mathbf{z}^{5}$, requiring no changes to the framework.
(c) Details of our module applied to $\mathbf{z}^{5}$. The input neurons are first analyzed using our metric. Once monosemantic neurons are identified, they are inhibited using RD. The resulting processed layer has the same shape as the input.}
\label{fig:module}
\end{center}
\end{figure*}

However, neither of the two solutions can achieve the first goal (i.e., training $f_1$ to inhibit the activation of $z$ for a single feature). To be more specific, method (a) wastes all the gradient for $f_1$ in generating $z$, preventing the related parameters from being updated. Method (b) outputs $\mathcal{V}(z')=\Bar{z}$ by compensating for the gap. As the model obtains a good result with $z$, by calculating the gradient, $f_1$ will be updated to push its output value from $\Bar{z}$ to $z$, expressed as $\Bar{z}+\delta(z-\Bar{z})$, where $\delta>0$ up to the learning rate (Figure~\ref{fig:RD}(i)). Ironically, the actual output of $z$ is updated to $z+\delta(z-\Bar{z})$, deviating from $\Bar{z}$ by a factor of $1+\delta$:
$$\frac{z+\delta(z-\Bar{z})-\Bar{z}}{z-\Bar{z}}=1+\delta.$$ 
Therefore, these two simple solutions either contribute nothing to the deactivation of $z$ or even enhance its activation (Figure~\ref{fig:RD}(g,k)).

After delving deeper into the literature, it is possible that previous researchers also recognized the negative effects of monosemanticity, but they may have discontinued their efforts after obtaining bad results from implementing these naive solutions.

\subsubsection{Reversed Deactivation}
\label{sssec:reverse}

To address the aforementioned problems, we propose our method called Reversed Deactivation (RD for short). Following the above-mentioned definitions, we replace the original $z$ with $z'$ (Figure~\ref{fig:RD}(f)):
\begin{equation}
z'=-z+(\Bar{z}+z)_{ng}.
\end{equation}
Similar to baselines, RD also ensures the second goal of decreasing the dependence of output $\mathbf{o}$ on the activation of $z$. To be more specific, $f_2$ receives a value $\mathcal{V}(z')=\mathcal{V}({-z+(\overline{z}+z)_{ng}})=\overline{z}$, which provides little information and requires $f_2$ to learn the information from other neurons in $\mathbf{z}$ (Figure~\ref{fig:RD}(h)).

Besides, RD can inhibit the activation level of $z$ when given $\mathbf{x}$. In short, after calculating the gradient, $f_1$ will update the trainable parameter to push its output value from $\Bar{z}$ to $z$ (Figure~\ref{fig:RD}(i)). Different from method (b), the gradient on $z$ is reversed. To achieve the same shifted value $\Bar{z}+\delta(z-\Bar{z})$ mentioned above, an insight into the update can be expressed as:
\begin{align}
    \mathcal{V}(-{z}+(\Bar{z}+z)_{ng})&=\Bar{z}\\
\rightarrow\mathcal{V}(-{(z-\delta(z-\Bar{z}))}+(\Bar{z}+z)_{ng})&=\Bar{z}+\delta(z-\Bar{z})
\end{align}
The output of $f_1$ without post deactivation (i.e., $z-\delta(z-\Bar{z})$) is closer to $\Bar{z}$ with a factor $1-\delta$:
$$\frac{z-\delta(z-\Bar{z})-\Bar{z}}{z-\Bar{z}}=1-\delta,$$
which achieves deactivation (Figure~\ref{fig:RD}(j)). The formal and detailed theory is presented in the following lemma.
\begin{lemma}
\label{lemma:deact}
Given a trained model $f$ with a continuous second derivative and a Lipschitz continuous gradient, where $f$ achieves a desired output $\mathbf{o}$ with minimal loss $\mathcal{L}(\mathbf{o})$, in which $\textbf{o}=f(\mathbf{x})=f_2(f_1(\mathbf{x}), \mathbf{x})=f_2(\mathbf{z}, \mathbf{x})$ for input $\mathbf{x}$ based on its monosemantic neuron $z$ in layer $\mathbf{z}$, suppose that $\mathcal{L}(f_2(\cdot))$ monotonically increases with $|z'-z|$ for any other value $z'$ that replaces $z$. Then, with a sufficiently small learning rate $l$, by updating the model $f$ with gradient descent based on the neuron processed by the RD method, the activation of $z$ on input $\mathbf{x}$ can be inhibited.
\end{lemma}
The proof is given in Appendix~\ref{appx:proof:deact}.

Additionally, we conduct validation experiments to compare the optimization outputs based on different methods in subsection~\ref{ssec:exp:naive}. The results are consistent with the theory and the analysis of naive methods and RD, showing that RD is effective in inhibiting monosemanticity.

\subsection{Flexible Plug-in Module}
\label{subsec:method:module}
In this subsection, we demonstrate the implementation of our method, which can be inserted after any neuron layer to inhibit its monosemanticity for emergence induction. We name our module \textbf{M}onosemanticity-based \textbf{Eme}rgence \textbf{L}earning (MEmeL for short) and outline the advantages of MEmeL:
\begin{itemize}[leftmargin=*]
    \item Adaptivity: Our module is compatible with any design of framework, and no structural changes are needed after inserting it.
    \item Light weight: No additional trainable parameters are introduced. 
\end{itemize}

The details of MEmeL are displayed in Figure~\ref{fig:module}. We present a general framework of a neural network in Figure~\ref{fig:module}(a). The output $\mathbf{o}$ is derived from $\mathbf{x}$ based on the layers of neurons $\{\mathbf{z}\}$. Yellow arrows indicate the reliance of neuron layers on each other. Our method can be applied to any layer of neurons, such as $\mathbf{z}^3$ and $\mathbf{z}^5$ in Figure~\ref{fig:module}(b), where the original neurons are adjusted by our modules to inhibit monosemanticity.

Taking $\mathbf{z}^5$ as an example (Figure~\ref{fig:module}(c)), we first detect monosemantic neurons using our MS metric, as introduced in subsection~\ref{subsec:method:detect}. After identifying these neurons (colored red), we apply our Reverse Deactivation method, as described in subsection~\ref{subsec:method:inhibit}, to each of them.

For adaptability, our module (i) does not require a specific input format and (ii) outputs $(\mathbf{z}^5)'$ in the same shape as the input $\mathbf{z}^5$, ensuring format consistency during data propagation. Thus, no adjustments to the framework are needed to apply our module.

For lightweightness, neither of our two methods introduces any trainable parameters. Additionally, MEmeL focuses on presenting the idea of functionality \textbf{induction}. The methods we propose do not directly decrease monosemanticity during training, but instead induce the model to do so, supported by theoretical analysis. Once we have a well-trained model, the performance is expected to be robust even without induction. Therefore, during testing, the module can be directly removed without incurring any inference overhead. The results of the validation experiments also support our analysis of removing MEmeL during testing. (see Appendix~\ref{appx:exp:rvm}).

\begin{table*}[!t]
\caption{Results on GLUE Test datasets. 
We follow the setting of BERT to demonstrate results on 8 datasets and calculate the average score. The scores are F1 scores for QQP and MRPC, Spearman correlations for STS-B, and accuracy scores for the other tasks. All metrics are the larger the better with best results in \textbf{bold} font.}
\vspace{-10px}
\setlength\tabcolsep{10pt}
\label{table:result:glue}
\begin{center}
\begin{tabular}{lccccccccc}
\toprule
\bf Model & \bf MNLI-(M/MM) & \bf QQP & \bf QNLI & \bf SST-2 & \bf CoLA & \bf STS-B& \bf MRPC& \bf RTE & \bf Average \\
\midrule
\bf Original    & 84.6/83.4& 71.2& 90.5&93.5&52.1&85.8&88.9&66.4&79.6 \\
\bf Naive (a)    & 84.3/83.6& \textbf{71.7}& 90.6&\textbf{93.8}&52.1&85.8&88.2&66.4&79.6 \\
\bf Naive (b)   & 84.7/\textbf{84.1}& 71.6& 90.6& 93.6&51.8&{86.5}&87.2&68.0&79.8 \\ \midrule
\bf MEmeL & \textbf{84.8}/83.9&\textbf{71.7}&90.9&93.6&{54.5}&\textbf{86.6}&{87.6}&\textbf{68.2}&{80.2}\\
\bf MEmeL-Tune & \textbf{84.8}/83.9&\textbf{71.7}&\textbf{91.2}&93.7&\textbf{55.7}&\textbf{86.6}&\textbf{89.0}&\textbf{68.2}&\textbf{80.5}\\
\bottomrule
\end{tabular}
\end{center}
\vspace{-5px}
\end{table*}

The detailed algorithm is provided in the Appendix~\ref{appx:module}. We also apply two simple tricks for implementation: late start and variance compensation, to avoid unstable value fluctuations during the cold start.

Last but not least, we emphasize the proposition of the pipeline for emergence-based learning. Researchers can focus on separable directions: (1) improving the metric to better detect factors related to scale change, and (2) designing factor-oriented solutions for better performance.

\section{Experiments}
\label{sec:exp}

In this section, we first introduce the basic experimental setup in subsection~\ref{subsec:exp:set},
then present the main empirical studies in subsection~\ref{subsec:exp:main}.
We show a case study that the proposed Reverse Deactivation 
is more powerful in inhibition monosematicity than the naive methods in subsection~\ref{ssec:exp:naive}.
Furthermore, we discuss the potentials and limitations of MEmeL in subsection~\ref{ssec:exp:limit}.
Note the experiment about the impact of removing MEmeL during test is provided in Appendix~\ref{appx:exp:rvm} due to the limited space.

\subsection{Experimental Setup}
\label{subsec:exp:set}

\begin{table}[t]
\caption{Results on ImageNet-1k dataset, where 3 sizes of Swin-Transformer pretrained on ImageNet-22k are used as backbones. 
The metric used is top-1 accuracy, where a higher value indicates better performance. 
The best results are indicated in \textbf{bold} font.}
\label{table:result:image}
\setlength\tabcolsep{10pt}
\begin{center}
\begin{tabular}{lccc}
\toprule
\bf Model & \bf Swin-T & \bf Swin-S & \bf Swin-B \\
Size & 28M&50M&88M\\
\midrule
\bf Original    & 80.9& 83.2& 85.1\\
\bf Naive (a) &81.0 &83.4&84.6\\
\bf Naive (b) &81.0 &83.4&85.1\\ \midrule
\bf MEmeL & \textbf{81.1}&83.4&85.1\\
\bf MEmeL-Tune & \textbf{81.1}&\textbf{83.5}&\textbf{85.2}\\
\bottomrule
\end{tabular}
\end{center}
\end{table}

\begin{table}[t]
\caption{Results on HKO-7 dataset. We initially trained a ConvGRU model for 20k steps to create the base model. 
The metrics used are B-MSE and B-MAE, where a smaller value indicates better performance. The best results are in \textbf{bold} fonts. We repeated each experiment three times and reported the average scores.}
\setlength\tabcolsep{10pt}
\label{table:result:rain}
\begin{center}
\begin{tabular}{lcc}
\toprule
\bf Model  & \bf B-MAE& \bf B-MSE \\
\midrule
\bf Original&1003.41 &309.96\\
\bf Naive (a)&1003.56 &309.83\\
\bf Naive (b)&1003.40 &310.13\\ \midrule
\bf MEmeL &1003.25&309.94\\
\bf MEmeL-Tune &\textbf{998.81}&\textbf{298.16}\\
\bottomrule
\end{tabular}
\end{center}
\end{table}

\begin{table*}[t!]
\caption{Validation experiments conducted on the Swin-B model. We record the Decrease Ratios and Update Scales of 10k neurons. The model that utilizes our Reverse Deactivation method is compared with those using two Naive methods and the original Swin-B.}
\label{table:result:valemel}
\vspace{-7px}
\setlength\tabcolsep{10pt}
\begin{center}
\begin{tabular}{lccc|c}
\toprule
\bf Methods & \bf Original & \bf Naive (a) & \bf Naive (b)& \bf Reverse Deactivation \\
\midrule
\bf Average Decrease Ratio&0.003\%&-0.017\%&-0.044\% &\textbf{0.013\%}\\
\bf Average Total Update Ratio &0.052\%&0.118\%&0.161\%& 0.189\%\\
\bottomrule
\end{tabular}
\end{center}
\vspace{-5px}
\end{table*}

\subsubsection{Data Sets, Base Models, and Tasks}
To validate the conjecture of monosemantic neurons, our inhibition method, and corresponding theories,
we apply MEmel to milestone models such as Bert, Transformer, and ConvRNN on various tasks (e.g., language, image, and physics simulation), respectively.

\begin{itemize}[leftmargin=*]
    \item Language Task. We apply MEmeL to the BERT (Pre-training of Deep Bidirectional Transformers) model \cite{BERT} on the GLUE (General Language Understanding Evaluation) benchmark \cite{GLUE}. BERT utilizes a transformer architecture and is pretrained on a large corpus of text data. It excels at capturing context and generating high-quality representations of words and sentences. 
    GLUE serves as a benchmark for natural language understanding tasks, encompassing various tasks such as natural language inference, sentiment analysis, and similarity analysis.

    \item Image Task. We conduct experiments on the Swin-Transformer model \cite{SWIN} and ImageNet dataset \cite{ImageNet}. The Swin-Transformer is a transformer model that uses a hierarchical structure and shift-based windows to capture cross-window connections. Our experiment follows their approach, which involves fine-tuning the Swin-Transformer on ImageNet-1K using checkpoints pretrained on ImageNet-22K. ImageNet-1K is a classification benchmark with 1,000 classes, consisting of 1.28 million training images and 50,000 validation images. 
    
    \item Physics Simulation Task. We apply our MEmeL to the ConvGRU model
    on the HKO-7 dataset~\cite{HKO}, 
    which forecasts precipitation based on images of radar echoes~\cite{ConvLSTM}. 
    ConvGRU is a lightweight module that belongs to the ConvRNN class, a milestone structure that combines CNN and RNN to capture spatiotemporal correlations simultaneously. 
    HKO is a large-scale benchmark dataset from the Hong Kong Observatory (HKO), providing high-resolution radar images spanning multiple years.
    Generally, precipitation forecasting is a challenging task, for both theory-driven and data-driven approaches, as it exhibits complex chaotic dynamics \cite{naturerainchaos, predrnn}. 
\end{itemize}

The configuration of above base models generally follow their original settings (see more details in Appendix~\ref{appx:exp:config}).
The metrics can be found in Appendix~\ref{appx:metric}.
We emphasize that our method is a general module applicable to models of any scale. After applying our module, the effectiveness of MEmeL can be evaluated by comparing it with the original model.

\subsubsection{Hyper-parameter Setting}
At the beginning of the experiment design, we tend to focus on the performance improvement brought by introducing reverse deactivation into milestone models.
To fairly validate the influence of the proposed MEmeL with other deactivation approaches (Naive (a) and (b)), we deactivate the neuron with \textbf{only top-$1$} MS in each batch (recorded as $\rm{MEmeL}$ vs. $\rm{Naive (a)}$ and $\rm{(b)}$).
During further discussion, people raised interest about the potential of tuning and we also display the results with tuned hyper-parameters (recorded as
$\rm{MEmeL-Tune}$). See more details in Appendix~\ref{appx:exp:search}.

\subsubsection{Implementation}
Our experiment is conducted on 4 V100 GPU cards. All the codes are implemented in PyTorch \cite{pytorch}, which are available through the link \href{https://github.com/dominatorX/MEmeL-code}{https://github.com/dominatorX/MEmeL-code}. 

\subsection{Main Experiment Result}
\label{subsec:exp:main}

Table~\ref{table:result:glue},  Table~\ref{table:result:image}, and  Table~\ref{table:result:rain} demonstrate the performance of MEmeL incorporated with BERT, Swin-Transformer, and ConvGRU, respectively.
The ``Original'' method denotes the raw model,
``Naive (a)'' and ``Naive (b)'' are corresponding to the original model incorporated with naive inhibition methods (see details in Equation~\eqref{eq:naive}).
By comparing the original method and the method enhanced with MEmeL, we can see that the MEmeL (especially MEmeL-Tune) achieves better or comparable results on different tasks, different data sets, and different base models. We also conduct the paired t-test on all three datasets to verify that the improvements are statistically significant (see Appendix~\ref{appx:exp:stat}).
This validates that neural networks can achieve better results by deactivating monosemantic neurons.
By comparing Naive (a), (b) with MEmeL (and MEmeL-Tune), we can easily observe that naive deactivation methods may intensify the monosemanticity of neural networks, which may lead to inferior performance improvement as discussed in subsection~\ref{sssec:naive}.

\subsection{Case Study on Deactivating Monosemantic Neurons}
\label{ssec:exp:naive}

To validate that the monosemanticity is indeed inhibited by our reverse deactivation,
we conducted experiments on the ImageNet-1k dataset using the Swin-Transformer model as shown in Table~\ref{table:result:valemel}.
We collected 10k samples for 4 different settings, where the modification of $z$ was done using Original (no modification), and two naive methods (a) and (b), and Reverse Deactivation in subsection~\ref{sssec:reverse}.

We feed inputs $\mathbf{x}$ to the model again to check how $z$ is optimized after updating the model with $\mathbf{x}$.
The new value is denoted as $z'$. 
Then, the Decrease Ratio represents how the monosemanticity is decreased upon $z$:
$\textit{Decrease Ratio}=\left(1-\phi(z')/\phi(z)\right)\times 100\%$.
Without any modification, monosemanticity showed a slight decrease with a small positive average decrease ratio (0.003\%) in the original model. 
Naive methods (a) and (b) intensify monosemanticity since their decrease ratios are negative, which is consistent with our analysis in Sec.~\ref{sssec:naive}.
On the contrary, reverse deactivation has the most significant impact on decreasing monosemanticity (0.013\%). 
But the value is relatively small.
That is mainly because the small learning rate ($2\times10^{-6}$) usually leads to a small update step in training.
To provide a clear illustration, we further define the Total Update Ratio as $\textit{Total Update Ratio}=\left|z'/z-1\right|\times 100\%$.
The scale of modification on $z$ and on $\phi(\cdot)$ is compatible (e.g., 0.013\% versus 0.189\% for RD). 
Thus, the small scale of the Decrease Ratio is due to the small learning rate, indicating a stable impact on the inhibition of monosemanticity using our method.

\subsection{Potential and Limitation of MEmeL}
\label{ssec:exp:limit}

According to our hypothesis, MEmeL induces the model to accumulate general and abstract functionality instead of monosemanticity for a specific task, which is consistent with the goal of per-taining. 
Although MEmeL achieves good results during fine-tuning (demonstrated at Main Experiments in subsection~\ref{subsec:exp:main}), the improvement is expected to be even greater when it is applied to the pre-training phase.

The main obstacle is the high computational resource cost required for pre-taining. 
Currently, we have only completed validation on relatively small precipitation forecasting datasets. 
In Table~\ref{table:result:rd2origin}, the model pre-trained with MEmeL (P-MEmeL) outperforms the one without it (P-Original). 
Based on these two models, the same finetuning process is conducted without MEmeL. 
Using MEmeL during pretraining improves B-MAE by 0.18\% and B-MSE by 1.29\% (T-MEmeL). In contrast, using MEmeL during finetuning improves B-MAE by 0.02\% B-MSE by 0.01\% (MEmeL in Table~\ref{table:result:image}). The results validate our hypothesis.

\begin{table}[!t]
\caption{Validation experiments conducted on the HKO-7 dataset. In addition, we pretrain a ConvGRU model for 20k steps using MEmeL, labeled as ``P-MEmeL''. The ``P-Original'' model is pretrained based on the original model and is used in the main experiment. Based on these two models, we perform finetuning for 2k steps using the original model, labeled as ``T-MEmeL'' and ``T-Original'', respectively. The metrics used for evaluation were B-MSE and B-MAE, where a smaller value indicates better performance. The best results are shown in bold fonts.}
\label{table:result:rd2origin}
\setlength\tabcolsep{10pt}
\begin{center}
\begin{tabular}{lcc}
\toprule
\bf Model & \bf B-MAE & \bf B-MSE \\
\midrule
P-Original&1004.25&311.54\\
P-MEmeL &\textbf{1000.98}&\textbf{306.67}\\
\midrule
T-Original &{1003.41}&{309.96}\\
T-MEmeL &\textbf{1001.56}&\textbf{305.96}\\
\bottomrule
\end{tabular}
\end{center}
\end{table}

\section{Conclusion}
\label{sec:conclusion}

Different from the literature, we hypothesize a key factor that highly promote the performance of large neural networks: the reduction of monosemantic neurons.
There is no unified quantitative evaluation metric and simply banning monosemantic neurons does not promote polysemanticity in neural networks.
Therefore, we propose to learn from emergence and present a study on proactively inhibiting the monosemantic neurons in this paper.
More specifically, 
we first propose a new metric to measure the monosemanticity of neurons with the guarantee of efficiency for online computation, then introduce a theoretically supported method to suppress monosemantic neurons and proactively promote the ratios of polysemantic neurons in training neural networks.
We validate our conjecture that monosemanticity brings about performance change at different model scales on a variety of neural networks and benchmark datasets
in different areas, including language, image, and physics simulation tasks.
Further experiments validate our analysis and theory regarding the inhibition of monosemanticity.

Unfortunately, extending this research to very large data sets or models (e.g., large language models) is appealing yet impossible for research departments due to limited resources. 
Therefore, we are trying to find ways to extend this paper to extremely large models trained on large data sets as future work.

\begin{acks}
Lei Chen’s work is partially supported by National Key Research and Development Program of China Grant No. 2023YFF0725100, National Science Foundation of China (NSFC) under Grant No. U22B2060, the Hong Kong RGC GRF Project 16213620, RIF Project R6020-19, AOE Project AoE/E-603/18, Theme-based project TRS T41-603/20R, CRF Project C2004-21G, Guangdong Province Science and Technology Plan Project 2023A0505030011, Hong Kong ITC ITF grants MHX/078/21 and PRP/004/22FX, Zhujiang scholar program 2021JC02X170, Microsoft Research Asia Collaborative Research Grant and HKUST-Webank joint research lab grants.
\end{acks}




{\bibliographystyle{ACM-Reference-Format}
\bibliography{EmeL}}


\appendix

\begin{figure*}[th!]
\begin{center}
\centerline{\includegraphics[width=\linewidth]{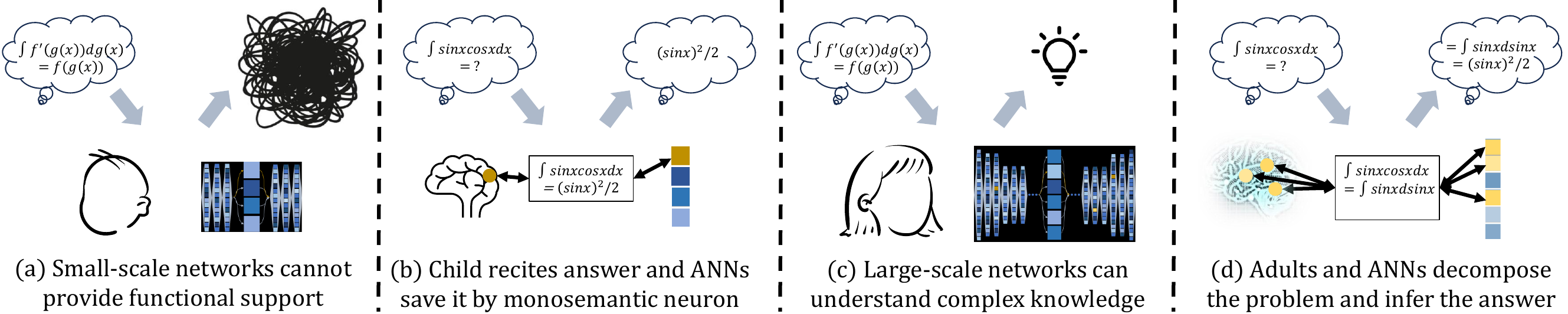}}
\caption{An analogy example to demonstrate our motivation:
(a) ANNs could be similar to biological brains. Small-scale networks (both biological and artificial) cannot support complex functionality.
(b) To solve a difficult problem that requires complex reasoning, pupils rely on rote memorization, while ANNs store QA pairs using monosemantic neurons as key-value pairs.
(c) Large-scale networks can learn and master solving skills related to integration.
(d) Adults and large-scale ANNs are capable of decomposing features and inferring answers using complex neuron circuits. This ability reduces reliance on rote memory and monosemanticity.}
\label{fig:analogy}
\end{center}
\end{figure*}

\section{An Example Analogy that Helps Illustrate Our Motivation}
\label{appx:example2}
We display an example with neuroscience analogy to better illustrate our motivation. For more relationships with neuroscience, see discussions in subsection.~\ref{appx:dis:bio}.

From the perspective of scale differences, when the model is small, the limited number of neurons makes it difficult to abstract and analyze the input, and to provide functional support for integrating a large amount of scattered information (Figure~\ref{fig:analogy}(a)). To achieve good results, neurons tend to specialize in certain features, resulting in stronger monosemanticity (Figure~\ref{fig:analogy}(b)).

We can analogize a small neural network to a developing brain of a pupil, memorizing question-answer pairs. Constructing monosemantic neurons is easier to achieve good results in complex tasks (such as accurately answering ten complex integration problems), compared to expecting the pupil to understand the solving steps and reason the results by themselves. (Not to mention that providing datasets with solving steps is very rare in neural network training, and small models struggle to automatically acquire reasoning abilities based solely on input-output pairs through simple gradient descent).

Similarly, we can analogize large models to adults, allowing them to learn and master solving skills related to integration (Figure~\ref{fig:analogy}(c)). This is because adults have accumulated knowledge and have a more mature brain development, similar to the huge training volume and parameter size of large models. In this process, adults gradually reduce the reliance on rote memorization and improve their reasoning abilities for new problems (Figure~\ref{fig:analogy}(d)). This can explain why large models have fewer monosemantic neurons and stronger generalization abilities in complex tasks.

To utilize this in model optimization, one issue is that existing large models are passive in decreasing monosemanticity, which can be likened to students lacking systematic education. Large models gradually reduce monosemanticity by accumulating training data, similar to students acquiring skills through extensive learning rather than rote memorization, and good guidance can help students improve faster. This leads us to our question: Can we proactively guide the reduction of monosemantic neurons to improve model performance?

Ideally, models will improve regardless of their scales. This is because large models also have the potential to decrease monosemanticity, resulting in better results.

\section{Related Works}
\label{sec:related}

\subsection{Artificial Neural Networks and the Increasing Scale}
Since \citet{neuronconcept} first modeled a simple neural network, ANN has been investigated under a scale of several layers in the last century \cite{zip, cnn}. With improvements in hardware, deeper models can be supported. AlexNet, which uses 8 layers for image classification, achieved excellent performance in the 2012 ImageNet challenge \cite{alex}. Later on, deeper models such as Inception and VGG increased the scale to tens and hundreds of layers \cite{inception, vgg}. The design of critical modules, such as ResNet, also plays a crucial role in ensuring the stability of model training when increasing the depth \cite{resnet}. 

After the Transformer was proposed and validated as effective in various areas \cite{attention}, both its scale and performance have seen significant growth over the years \cite{BERT, SWIN}. In recent years, architectures based on the Transformer have achieved great success at extremely large scales \cite{GPT3, LLAMA2}. 
Investigating the underlying effective properties created by the increase in scale is a highly demanded research direction. 

\subsection{Mechanistic Interpretability}
\label{subsec:related:mono}
With the improved performance of neural networks, their black-box nature raises more questions than it answers. To gain a better understanding of and diagnose neural networks, researchers seek mechanistic interpretability \cite{curcuitsurvey}. They study individual components to understand their functionality and usage, such as neurons for identifying dogs and cars \cite{circuits}. As Transformer models demonstrate their superiority in various domains, there is an increasing focus on their interpretability. \citet{keyvalue} propose that the feed-forward layers of Transformers function as key-value pairs. \citet{embeddingkv} extend this mapping to the embedding space. Although the complexity greatly increases with larger models, the success of these models attracts researchers who strive to find interpretability in the vast sea of neurons \cite{sparseprobe,bricken2023towards,GPT42GPT2}. In their work, \citet{softmax} introduce the softmax linear unit to create monosemantic models. Additionally, \citet{sparseprobe} modify the sparse probe by using multiple heads to classify neurons with decreasing monosemanticity. \citet{bricken2023towards} construct more powerful and complex probes to construct mappings between neurons and features. While the desire to decompose and understand everything is appealing (e.g., achieving monosententicity), it may conflict with the progress of intelligence.

\subsection{Information Bottleneck}
As an important method to explain the deep learning mechanism, the Information bottleneck (IB) provides a compression view for deep learning and emphasizes forgetting together with information retaining \cite{IB}. For large-scale models, in addition to understanding how much and what information the model retains, understanding how the model manipulates the retained information is also important. Our method encourages the model to organize information more abstractly and distributed into multiple neurons, which is positive to performance supported by existing analysis with biological analogies.

For example, we hope to punish \textbf{each} neuron if it only provides information for a specific feature, rather than considering the total information that \textbf{a layer of} neurons retains and forgets. In other words, we emphasize the way neurons \textbf{manipulate} the retained information (distributed or one-to-one), while IB emphasizes the total \textbf{amount} of information that different layers of neurons record at different training stages \cite{IBlearn}.

As the scale of neural networks continues to grow, effective information processing and reasoning have become as important a need as information compression. In addition to what information should be squeezed through the bottleneck, how to process the squeezed information is also important and is what our paper concerns \cite{newIB}. Our method is a promising direction to enrich deep learning understanding together with IB, especially when large model application requires more and more powerful reasoning ability. 

\subsection{Debate on the Existence of Emergence}
Note that \citet{DBLP:journals/corr/abs-2304-15004} states that ``emergent abilities appear due to the researcher’s choice of metric rather than due to fundamental changes in models with scale''. Here we point out that their finding does not diminish the value of our finding, but instead partially coincides with our idea: 
\begin{itemize}[leftmargin=*]
    \item Though evaluation metrics can be smooth and well-designed, models are improved based on training data. However, solving hard and realistic problems via advanced AI involves more and more data with poorly labeled or even without labels. Emergence learning is demanded to find and boost the underlying ability accumulation of models, which diminishes the correctness of individual answers and focuses on the potential knowledge learning brought by scale changes.
    \item To observe the accumulated ability of the model on these hard problems, we need smooth and mild metrics. Such metrics may not be available for challenging problems in the future, where the research would be like roaming at deep night. Factors discovered through Emergence Learning can help validate the potential improvement of models and enlighten the darkness.
\end{itemize}

\begin{algorithm}[!t]
   \caption{Monosementicity Scale Computing with Needed Variables}
   \label{alg:metric}
\begin{algorithmic}
   \STATE {\bfseries Input:} new batch of values $\left\{\mathbf{z}^{[m+1]},\mathbf{z}^{[m+2]},\cdots,\mathbf{z}^{[m+b]}\right\}$ of the neuron.
   \STATE {\bfseries Local Variables:} forget step $n_f$, current step $c_t$, current sample mean $\mu_m$ and variance $\upsilon_m$ 
   \STATE Calculate the MS for each input value $z\in \mathbf{z}$: $\phi(z^{[m+i]}) = (z^{[m+i]}-\mu_m)^2/\upsilon_m$ for $i\in [1,b]$.
   \STATE Calculate $\mu_b'=\frac{1}{b}\sum_{i=1}^bz^{[m+i]}$ and $\upsilon'_b=\frac{1}{b}\sum_{i=1}^b(z^{[m+i]}-\mu'_b)^2$.
   \IF{$c_t < n_f$}
   \STATE $\mu_{m+b}=\frac{m\mu_m+b\mu'_b}{m+b}, $
   $\upsilon_{m+b}=\frac{mb(\mu_m-\mu_b')^2}{(m+b-1)(m+b)}+\frac{b\upsilon'_b+(m-1)\upsilon_m}{m+b-1}$
   \ELSE
   \STATE $\mu_{m+b}=\frac{n_f\mu_m+\mu'_b}{n_f+1}, $
   $\upsilon_{m+b}=\frac{n_f(\mu_m-\mu_b')^2}{(n_f+1-1/b)(n_f+1)}+\frac{\upsilon'_b+(n_f-1/b)\upsilon_m}{n_f+1-1/b}$
   \ENDIF
   \STATE {\bfseries RETURN:} Monosementicity Scale of inputs $\phi({\mathbf{z}})$
\end{algorithmic}
\end{algorithm}

\begin{algorithm}[!t]
   \caption{Monosementicity-based Emergence Learning}
   \label{alg:module}
\begin{algorithmic}
   \STATE {\bfseries Input:} Values of $n$ neurons with batchsize $b$: $\mathbf{z}=\{\{z_i^{[j]}\}_{j=1}^b\}_{i=1}^n$.
   \STATE {\bfseries Local Variables:} late start step $l_s$, current step $c_t$, current sample mean $\{{\mu_i}_m\}_{i=1}^n$ and variance $\{{\upsilon_i}_m\}_{i=1}^n$ 
   \IF{$c_t < l_s$}
   \STATE Calculate the MS for each input value 
   $$\phi(z_i^{[m+j]}) = \frac{(z_i^{[m+j]}-{\mu_i}_m)^2}{{\upsilon_i}_m+\epsilon \sum_{k=1}^{n}{\upsilon_k}_m/n} \text{ for } i\in[1, n], j\in [1,b].$$
   
   \STATE Select values with high $\phi$, adjust each of them according to MD. Replacing the original ones in $\mathbf{z}$ to form output $\mathbf{z}'$.
   \ELSE
   \STATE $\mathbf{z}'=\mathbf{z}$
   \ENDIF
   \STATE Update sample mean $\{{\mu_i}_m\}_{i=1}^n$ and variance $\{{\upsilon_i}_m\}_{i=1}^n$ using Algorithm~\ref{alg:metric}.
   \STATE {\bfseries RETURN:} Adjusted layer of neurons $\mathbf{z}'$
\end{algorithmic}
\end{algorithm}

\section{Implementation Details}
\subsection{Implementation Details of MS $\phi$}
\label{appx:metric}
During training, the model is continuously updated. However, samples from the early stages become outdated, and more importance should be given to the new samples. To address this, we introduce a forget step ($n_f$) and keep track of the current training steps ($c_t$). Once $c_t\geq n_f$, we update the sample mean and variance by replacing the number of samples from $m$ to $b\cdot n_f$. This reduces the influence of previous samples. Both variable updating and mean and variance computation have a complexity of $O(1)$.

\subsection{Implementation Details of MEmeL}
\label{appx:module}
As described in Algorithm~\ref{alg:metric}, the statistical variables are calculated online. However, due to the limited number of samples at the beginning of training, the estimation can be unstable. Additionally, if the estimation of $S^2$ is extremely small, it may cause overflow during calculation. To improve the robustness of training, we introduce a late start step $l_s$ and a variance compensation in the denominator of the metric calculation in Equation~\eqref{eq:phi}. As outlined in Algorithm~\ref{alg:module}, when a batch of data arrives, we only update neurons with RD if the current step is greater than $l_s$. The calculation of MS is also adjusted by incorporating the mean of variances of other neurons, weighted by a small value $\epsilon$ to prevent overflow.

\section{Proof of Section~\ref{sec:method}}

\subsection{Proof of Lemma~\ref{lemma:phib}}
\label{appx:phib}

\begin{proof}
Recall that we define:
$$\mu_m=\frac{\sum_{i=1}^mz^{[i]}}{m},\mu_b'=\frac{\sum_{i=1}^bz^{[m+i]}}{b}.$$

We have:
\begin{align}
\mu_{m+b}&=\frac{\sum_{i=1}^{m+b}z^{[i]}}{m+b}\\
&=\frac{\sum_{i=1}^{m}z^{[i]}+\sum_{i=1}^{b}z^{[m+i]}}{m+b}\\
&=\frac{m\mu_m+b\mu'_b}{m+b}.\label{eq:proof:mu}
\end{align}

Besides, we define:
$$\upsilon_m=\frac{\sum_{i=1}^m(z^{[i]}-\mu_m)^2}{m-1}, \upsilon'_b=\frac{\sum_{i=1}^b(z^{[m+i]}-\mu'_b)^2}{b}.$$

Then we have:
{\tiny 
\begin{align}
\upsilon_{m+b}=&\frac{\sum_{i=1}^{m+b}(z^{[i]}-\mu_{m+b})^2}{m+b-1}=\frac{\sum_{i=1}^{m+b}(z^{[i]}-\frac{m\mu_m+b\mu'_b}{m+b})^2}{m+b-1}\\
=&\frac{\sum_{i=1}^{b}(z^{[m+i]}-\frac{m\mu_m+b\mu'_b}{m+b})^2+\sum_{i=1}^{m}(z^{[i]}-\frac{m\mu_m+b\mu'_b}{m+b})^2}{m+b-1}\\
=&\frac{\sum_{i=1}^{b}\left[z^{[m+i]}-\mu'_b-\frac{m(\mu_m-\mu'_b)}{m+b}\right]^2+\sum_{i=1}^{m}\left[z^{[i]}-\mu_m+\frac{b(\mu_m-\mu'_b)}{m+b}\right]^2}{m+b-1}\\
=&\frac{b\left(\frac{m(\mu_{m}-\mu_{b}^{\prime})}{m+b}\right)^{2}+\sum_{i=1}^{b}\left(z_{(i+m)}-\mu'_{b}\right)^{2}-\sum_{i=1}^{b}\left[2(z^{[m+i]}-\mu'_{b})\frac{m(\mu_{m}-\mu_{b}^{\prime})}{m+b}\right]}{m+b-1}\label{eq:proof:acc1}\\
&+\frac{m\left(\frac{b(\mu_{m}-\mu_{b}^{\prime})}{m+b}\right)^{2}+\sum_{i=1}^{m}\left(z^{[i]}-\mu_{m}\right)^{2}+\sum_{i=1}^{m}\left[2(z^{[i]}-\mu_{m})\frac{b(\mu_{m}-\mu_{b}^{\prime})}{m+b}\right]}{m+b-1}\label{eq:proof:acc2}\\
=&\frac{mb^{2}(\mu_{m}-\mu_{b}^{\prime})^{2}+bm^{2}(\mu_{m}-\mu_{b}^{\prime})^{2}}{(m+b-1)(m+b)^{2}}+\frac{\sum_{i=1}^{b}\left(z^{[m+i]}-\mu_{b}^{\prime}\right)^{2}-0+(m-1)\upsilon_{m}+0}{m+b-1}\\
=&\frac{mb(\mu_m-\mu_b')^2}{(m+b-1)(m+b)}+\frac{b\upsilon'_b+(m-1)\upsilon_m}{m+b-1},
\end{align}
}
where the third terms in Equation~\eqref{eq:proof:acc1} and Equation~\eqref{eq:proof:acc2} equal to zero as $\sum_{i=1}^{b}(z^{[m+i]}-\mu'_{b})=\sum_{i=1}^{b}(z^{[m+i]})-b\mu'_{b}=0, \sum_{i=1}^{m}(z^{[i]}-\mu_{m})=\sum_{i=1}^{m}(z^{[i]})-m\mu_m=0$ and other terms are constant factors.
\end{proof}

\subsection{Proof and Discussion of Lemma~\ref{lemma:deact}}
\label{appx:proof:deact}
\begin{proof}
Recall that based on Reversed Deactivation, our gradient flow originates from the following equations:
\begin{align}
\mathbf{o}'=&f_2(\mathbf{z}', \mathbf{x})=f_2([-z+(z+\Bar{z})_{ng}, \mathbf{z}^-, \mathbf{x}]),\label{eq:f2}\\
\mathbf{z}=&[z,\mathbf{z}^-]=f_1(\mathbf{x}),\label{eq:f1}
\end{align}
where $\cdot_{ng}$ denotes ``no gradient'' scalar and ${\mathbf{z}^-}\subset \mathbf{z}$ are the neurons in $\mathbf{z}$ except $z$.

Though a bit of abusing the symbol, to give a clearer representation for gradient update, we use $z$ to refer to the neuron, $z_a$ refers to its activation \emph{value} which leads to an ideal $\mathbf{o}$. $z_o=z_a$ as the value of $z$ before processed by RD, $z^*$ for the value updated after gradient descent. We still use $\Bar{z}$ as the mean value of $z$ and represent the difference between $z$ and $\Bar{z}$ as an updating variable $t=z-\Bar{z}$. Note that except $z$ and $t$, other definitions in this paragraph are untrainable scalars.

Here we focusing on $z-$related parts of Equation~\eqref{eq:f2} and \eqref{eq:f1}. We denote parameters of $f_1$ as $\theta\in \mathbf{W}$ and rewrite Equation~\eqref{eq:f1} for gradient calculation and parameter update:
\begin{align}
z=&g(\theta)+\Bar{z}\\
\Rightarrow t=&g(\theta),\label{eq:g1}
\end{align}
where $g$ is $z$-related functions of $f_1$. 

Here we fix other states and focus on the interaction between $z$, $\theta$, and loss $\mathcal{L}(\mathbf{o}')$. We can calculate the gradient for $\theta$:

\begin{align}
\nabla \mathcal{L}(\theta)=&\frac{\partial{\mathcal{L}(\mathbf{o}')}}{\partial \theta}\\ =& \frac{\partial{\mathcal{L}(f_2(\mathbf{z}'))}}{\partial t}\frac{\partial t}{\partial \theta}\\
=&\frac{\partial{\mathcal{L}(f_2(\mathbf{z}'))}}{\partial (z'-z_a)}\frac{\partial(z'-z_a)}{\partial t}\frac{\partial t}{\partial \theta}\\
=&-\nabla f(\delta)\nabla g(\theta),
\end{align}

where $\frac{\partial t}{\partial \theta}=\nabla g(\theta)$ as the gradient of Equation~\eqref{eq:g1} and $\frac{\partial(z'-z_a)}{\partial t}=\frac{\partial(-z+z_a+\Bar{z}-z_a)}{\partial (z-\Bar{z})}=-1$. $\nabla f(\delta)$ denotes $ \frac{\partial{\mathcal{L}(f_2(\mathbf{z}'))}}{\partial (z'-z_a)}.$ According to our assumption that $f(\delta)$ the monotonically increases with $|z'-z_a|$, $\nabla f(\delta)\geq0$ when $z'\geq z_a$ so that $z'-z_a=|z'-z_a|$, which means $\Bar{z}\geq z_a$ and $t\leq0$. Also, we can derive that $t\geq0$ when $\nabla f(\delta)\leq0$. 

With gradient descent of learning rate $l$, we will update parameter $\theta$ as:
\begin{equation}
    \theta^* \leftarrow \theta-l\nabla \mathcal{L}(\theta)=\theta+l\nabla f(\delta)\nabla g(\theta)
\end{equation}
By updating the parameter, we can express the updated $z$ with a variation on the multivariate Taylor expansion:
\begin{align}
z^*=&g(\theta^*)\\
=&g(\theta)+l\nabla g(\theta)^{T}\nabla f(\delta)\nabla g(\theta)+l^2\frac{1}{2}(\nabla \mathcal{L}(\theta))^{T}\nabla^2 g(\theta')(\nabla \mathcal{L}(\theta)),
\end{align}
where $\theta^*$ is the updated parameter and  $\theta'$ is between $\theta$ and $\theta^*$. 

When $\nabla f(\delta)\geq0$, we have
\begin{align}
z^*-z_o=&l\nabla g(\theta)^{T}\nabla f(\delta)\nabla g(\theta)+\frac{l^2}{2}(\nabla \mathcal{L}(\theta))^{T}\nabla^2 g(\theta')(\nabla \mathcal{L}(\theta)) \\
=&l(d_1+\frac{l}{2}d_2)\label{eq:output},
\end{align}
which is greater than 0 when $d_2\geq0, l>0$ or  $d_2<0, 0<l<\frac{2d_1}{-d_2}$, where $d_1=\nabla  f(\delta)||\nabla g(\theta)||^2$ and $d_2=(\nabla \mathcal{L}(\theta))^{T}\nabla^2 g(\theta')(\nabla \mathcal{L}(\theta))$.
As $t=z-\Bar{z}<0$ under $\nabla f(\delta)\geq0$, the activation value is smaller than the mean $\Bar{z}$. Note that after the update, $z^*>z_o$. With a small enough learning rate $l$ (i.e., $<\frac{2d_1}{-d_2}$), $z^*$ will be updated towards $\Bar{z}$ and deactivated (i.e., $\Bar{z}>z^*>z_o$) 

When $\nabla f(\delta)\leq0$, we have
\begin{align}
z^*-z_o=&l\nabla g(\theta)^{T}\nabla f(\delta)\nabla g(\theta)+\frac{l^2}{2}(\nabla \mathcal{L}(\theta))^{T}\nabla^2 g(\theta')(\nabla \mathcal{L}(\theta))\\
\leq&l\nabla f(\delta)||g(\theta)||^2+\frac{l^2L}{2}||\nabla g(\theta)||^2,\\
=&l(\nabla f(\delta)+l\frac{L}{2})||\nabla g(\theta)||^2  \label{eq:output2}
\end{align}
where Equation~\eqref{eq:output2} is valid as $g$ is the $z$-related part of model $f$ and has a Lipschitz continuous gradient. Equation~\eqref{eq:output2} is smaller than 0 when learning rate $0<l<(-\nabla  f(\delta))\frac{2}{L}$, which ensures that updated output $z^*$ is smaller than original activated value $z_o$, and thus close to $\Bar{z}$ as $t=z-\Bar{z}\geq0$.
Together with the case when $\nabla f(\delta)\geq0$, we show that $z$ is always pushed to mean $\Bar{z}$ and deactivated.

\end{proof}
We point out that the loss function $\mathcal{L}(\mathbf{o})$ does not always strictly increase with the deviation from $z$ (i.e., $|z'-z|$) in neural networks. This is due to the presence of local optima, where increasing deviation from $z$ may actually lead to a smaller loss. Fortunately, in the case of a monosemantic neuron, the output is strongly influenced by $z$ and exhibits less nonlinearity, aligning more closely with the assumption.

\begin{table*}[t!]
\caption{Validation results on GLUE Test datasets. The settings are the same with Table~\ref{table:result:glue}. All metrics are the larger the better with best results in \textbf{bold} font.}
\label{table:rvm:glue}
\begin{center}
\begin{tabular}{lccccccccc}
\toprule
Model & MNLI-(M/MM) & QQP & QNLI &SST-2 &CoLA &STS-B&MRPC&RTE&Average \\
\midrule
MEmeL & 
84.8/\textbf{83.9}&\textbf{71.7}&\textbf{90.9}&93.6&\textbf{54.5}&\textbf{86.6}&{87.6}&\textbf{68.2}&\textbf{80.2}\\
MEmeL-T & \textbf{84.9}/83.7&71.5&90.5&\textbf{93.9}&52.1&{85.9}&\textbf{88.7}&{68.1}&79.9\\
\bottomrule
\end{tabular}
\end{center}
\end{table*}

\begin{table}[t!]
\caption{Validation results on ImageNet-1k dataset. See Table~\ref{table:result:image} for detailed settings. The metric is the higher the better. The best results are indicated in  \textbf{bold} font.}
\label{table:rvm:image}
\begin{center}
\begin{tabular}{lccc}
\toprule
Model & Swin-T & Swin-S & Swin-B \\
Size & 28M&50M&88M\\
\midrule
MEmeL & \textbf{81.1}&83.4&\textbf{85.1}\\
MEmeL-T & \textbf{81.1}&\textbf{83.5}&\textbf{85.1}\\
\bottomrule
\end{tabular}
\end{center}
\vskip -0.1in
\end{table}

\begin{table}[t!]
\caption{Results on HKO-7 dataset. The settings are the same with Table~\ref{table:result:rain}. The metrics are the smaller the better with the best results in \textbf{bold} fonts.}
\label{table:rvm:rain}
\begin{center}
\begin{tabular}{lcc}
\toprule
Model & B-MAE & B-MSE \\
\midrule
MEmeL&1003.25 &309.94\\
MEmeL-T &\textbf{1002.46}&\textbf{309.52}\\
\bottomrule
\end{tabular}
\vspace{-5pt}
\end{center}
\end{table}

\section{Experiment Setting}
\label{appx:expset}

\subsection{Configuration of Base Models}
\label{appx:exp:config}

For the language task, we insert 4 MEmeL layers after the 4 middle Attention layers of BERT-base (12 in total). Since MEmeL does not introduce any additional parameters, we can directly use their checkpoint after pretraining to finetune. The finetuning setting is the same as BERT for fairness, where each task is trained with a batch size of 32 and 3 epochs over the data for all GLUE tasks. For each task, we selected the best fine-tuning learning rate ($l\in\{ 5e-5, 4e-5, 3e-5, 2e-5\}$) on the validation dataset (DEV). Baseline results are from Table~1 in BERT \cite{BERT}.

For the image classification task, we also insert 4 MEmeL layers after the 4 middle Attention layers for each model. The training is conducted on ImageNet-1k for 30 epochs using pretrained checkpoints on ImageNet-22k. The settings are directly adopted from their scripts, except for the number of GPU cards. Baseline results are collected by running checkpoints from the \href{https://github.com/microsoft/Swin-Transformer/}{github} of Swin-Transformer with some updates.  

For the precipitation task, we insert 6 MEmeL lay-
ers after the hidden states of each ConvGRU module. The training is conducted on the HKO-7 dataset \cite{HKO} of radar images with a granularity of $120\times120$, which are centered in Hong Kong. These radar images cover a timespan from 2009 to 2015 and are captured every 6 minutes. The original rainfall intensity is mapped to a scale of [0, 255], forming a regression task. The pretraining settings are the same as theirs, optimizing for 20k steps. Based on that, we conduct 2k more steps finetuning to obtain the final output.

\subsection{Metrics}
\label{appx:metric}

Here we list the metrics used in the experiments.
\begin{itemize}[leftmargin=*]
    \item The \textbf{Accuracy}: With the statistic of True Positive (TP), False Positive (FP), True Negative (TN), and False Negative (FN), Accuracy can be derived as $\frac{\text{TP+TN}}{\text{TP+TN+FN+FN}}$.
    \item The \textbf{F1 score}:
    $$F1 = \frac{{2 \cdot \text{{Precision}} \cdot \text{{Recall}}}}{{\text{{Precision}} + \text{{Recall}}}}, $$
    where Precision is calculated as $\frac{\text{TP}}{\text{TP+FP}}$ and Recall refers to $\frac{\text{TP}}{\text{TP+FN}}$.
    \item The \textbf{Spearman correlations}:
    $$\rho=\frac{\sum_i(x_i-\bar{x})(y_i-\bar{y})}{\sqrt{\sum_i(x_i-\bar{x})^2\sum_i(y_i-\bar{y})^2}},$$
    where $[x_1, x_2, \cdots]$ and $[y_1, y_2, \cdots]$ are two sequences.
    \item The \textbf{B-MSE} and  \textbf{B-MAE} (The balanced mean square error and balanced mean absolute error):
$$\text{B-MSE} = \frac{1}{D}\sum_{m\in M}w(m)(m-\hat{m})^2,$$
$$\text{B-MAE} = \frac{1}{D}\sum_{m\in M}w(m)|m-\hat{m}|,$$
where $M\in \mathrm{R}^{T\times H\times W}$ is the radar intensity map to be predicted with $D=T\times H\times W$ pixels. For each pixel with intensity $m$, $\hat{m}$ is the prediction output and $w(m)$ is a weight to enhance the performance on heavy rainfall:
$$w(m)=\begin{cases}\begin{array}{cc}1,&m<2\\2,&2\leqslant m<5\\5,&5\leqslant m<10\\10,&10\leqslant m<30\\30,&m\geqslant30\end{array}&\end{cases}$$

\end{itemize}

\subsection{Parameter Searching over Inhibition Level of Monosemanticity}
\label{appx:exp:search}

As mentioned above, to ensure fairness, we apply the minimal scale of inhibition, wherein only the most monosemantic neuron is selected and deactivated for each sample. However, we can also adjust the inhibition scale in each step by introducing a factor $\Delta$. By unifying the two naive methods and our reverse deactivation, we obtain the following function:
$$z'=-\Delta\cdot z+(\Bar{z}+\Delta\cdot z)_{ng}.$$
Our reverse deactivation (RD) is a special case when $\Delta=1$, while the two naive methods can be obtained with $\Delta=0$ and -1. It can be easily proven that for any $\Delta>0$, the modification can deactivate neuron $z$ according to Lemma~\ref{lemma:deact}, and a larger $\Delta$ corresponds to a more aggressive deactivation.
Therefore, we search for the number of neurons $n_n$ to deactivate in a single step and the scale of deactivation $\Delta$ for the best performance. We select the best model from the search space: $n_n\in \{1, 5, 10, 50, 100\}$ and $\Delta\in\{1, 5, 10, 50, 100\}$.

\section{More Experiments}
\subsection{The Impact of Removing MEmeL During Test}
\label{appx:exp:rvm}
Recall that in subsection~\ref{subsec:method:module}, we emphasize the lightness of our MEmeL. It aims to induce models to reduce monosemanticity during training and can be removed during testing. Here, we conduct experiments to enable MEmeL during testing, which are expected to have similar results compared to the results displayed in the main experiments (see Section \ref{subsec:exp:main}). These models are labeled with the postfix ``-T" (e.g., MEmeL-T for MEmeL as the base model) while keeping other experimental settings unchanged.

The results are displayed in Table~\ref{table:rvm:glue}, Table~\ref{table:rvm:image}, and Table~\ref{table:rvm:rain} for GLUE, ImageNet, and HKO-7 datasets, respectively. The best results are shown in bold fonts. The performance differences for the two settings are very close on all three tasks, indicating a stable performance when tested without the induction of MEmeL. Interestingly, on GLUE datasets, the performance when using MEmeL during testing is much more fluctuated. It would be valuable to study the dynamics of MEmeL in more depth as future work.

\begin{table}[t!]
\caption{The paired t-test on all 3 datasets. ``Mean-Original'' and `Mean-Ours'' refers to the mean scores of the two methods on each dataset. t-statistic>0 for GLUE and ImageNet as their metrics are the larger the better.
t-statistic<0 for HKO-7 as its metric is the smaller the better.}
\label{table:exp:stat}
\begin{center}
\begin{tabular}{lccc}
\toprule
Dataset & GLUE & ImageNet  &HKO-7 \\
\midrule
\bf{p-value}&\bf{0.02}	&\bf{0.07}&	\bf{0.06}\\
\midrule
t-statistic&	2.77&	3.46	&-2.38\\
Mean-Original	&79.6	&83.1&	656.69\\
Mean-Ours	&80.5	&83.3	&648.49\\
\bottomrule
\end{tabular}
\vspace{-5pt}
\end{center}
\end{table}

\subsection{Statistical Significance}
\label{appx:exp:stat}
To statistically verify the improvement from our method, we conduct the paired t-test on all three datasets. The results are given in Table.~\ref{table:exp:stat}. Each score is obtained from the results of the \textbf{Original} and \textbf{MEmeL-Tune}. All the datasets show a 90\% confidence level (p-value <0.1) supporting the hypothesis that our performance significantly differs from the baseline.

\section{Discussions}

\subsection{Biological View of Neuron Connection}
\label{appx:dis:bio}
Monosemanticity and polysemanticity are inherent mappings between neurons, where decreasing monosemanticity encourages the construction of complex mappings instead of exclusive connections. With a similar functionality, in neuroscience, ``synaptic'' enables information transmission between neurons \cite{Cellsynaptic, EffectSynaptic}.
In view of the evolution of the brain, the increase of neuron activity is positively correlated to the increase of the amount of synaptic \cite{IncreaseSynptic}. Besides, compared with chimpanzees, humans significantly have more synaptics for information transmission \cite{HumanSynptic}.

However, the connections of artificial neurons are fixed during design. In contrast, the connection relationships of biological neurons are observable, which can serve as a reference to study the monosemanticity of AI neurons in large-scale models.

\subsection{Brain-inspired ANN Design}
Researchers have developed various methods for incorporating biological knowledge into ANNs \cite{synapticneuronreview, brainannreview}. For example, to guide the design of ANNs, Spiking Neural Networks (SNNs) draw inspiration from the way neurons communicate in the brain, where information is encoded in the timing of spikes or action potentials \cite{bnnspike}. This spike-based communication enables SNNs to potentially achieve greater efficiency and better biological plausibility. In optimizing ANNs, Eligibility Propagation combines traditional error backpropagation with biologically plausible learning rules \cite{eprop}. This approach updates the model by calculating an eligibility trace for each synapse and measuring its contribution to the neuron.

These methods tend to help ANN by focusing on the functionality and mechanism of biological neurons, whereas EmeL emphasizes the importance of scale as an angle of improvement.

\subsection{Difference from Traditional Normalizations}
Here we demonstrate that our method is greatly different from normalization. 

Normalization calculates the mean and variance of \textbf{multiple neurons} in a layer given 1 single input. 

For our method, the mean and variance are computed for \textbf{each single neuron} with different inputs. 

The underlying reason is that a neuron is considered \textbf{activated} when its value of 1 input deviates from the mean among all the inputs (Our points). While for normalization, it should be smoothed if it deviates from other neurons.

For example, let’s consider 5 neurons in a layer $(h_1,h_2,h_3,h_4,h_5)$. Now assume that given the $i^{th}$ input $x^{(i)}$, the values of the 5 neurons are  $(h_1^{(i)},h_2^{(i)},h_3^{(i)},h_4^{(i)},h_5^{(i)})=$$(-1, 1, -1, 0, -2)$. 

\textbf{For normalization}: 

Calculating the mean and standard deviation over all these 5 values, which are both \textbf{scalar}:

$$\mu=\sum_{t=1}^5 h_t^{(i)}/5 = -0.6,$$

$$\sigma=\sqrt{(\sum_{t=1}^5 (h_t^{(i)}-\mu)/5)}=1.0.$$ 

Normalize the values of the 5 neurons: 

\begin{align}
Norm(h_1^{(i)},h_2^{(i)},h_3^{(i)},h_4^{(i)},h_5^{(i)})&=((-1, 1, -1, 0, -2)-\mu)/\sigma\nonumber\\
&=(-0.4, 1.7, -0.4,  0.6, -1.4)\nonumber
\end{align}

\textbf{For our deactivation}:

We will first dynamically calculate the mean and the standard deviation for each neuron based on all the previous inputs $\{x^{(t)}\}_{t=1}^{i-1}$, which would both be \textbf{5 values}:

historical mean: $(\mu_1,\mu_2,\mu_3,\mu_4,\mu_5) = (3, 0, -1, 1, -1),$

and standard deviation: $(\sigma_1,\sigma_2,\sigma_3,\sigma_4,\sigma_5)=(1, 1, 1, 1, 1).$

According to our metrics \textbf{MS}, neuron $h_1$ with value $h_1^{(i)}=-1$ and historical mean $\mu=3$ is the most \textbf{activated}. After our process, the output will become 

\begin{align}RD({h_1^{(i)}},h_2^{(i)},h_3^{(i)},h_4^{(i)},h_5^{(i)})&=((h_1^{(i)})',h_2^{(i)},h_3^{(i)},h_4^{(i)},h_5^{(i)})\nonumber\\&=(\mathbf{3}, 1, -1, 0, -2).\nonumber\end{align}

\subsection{Correctness of Equation~\ref{eq:followed}}
Equation~\ref{eq:followed} draws some misunderstanding that it should be written as $f_2(\mathbf{z})$. However, $f_2$ should have both $\mathbf{x}$ and $\mathbf{z}$ as inputs when the model is not sequential. For example, in Figure~\ref{fig:module}(b), when focusing on $\mathbf{z}^3$, $\mathbf{o} =f_2(\mathbf{z}^3, \mathbf{x})$ which is not only a function of $\mathbf{z}^3$ but also $\mathbf{z}^4(\mathbf{x})$.

\subsection{Discussion of Experimental Settings}
\label{appx:exp:select}
Currently, middle- to large-scale models typically follow a pretrain-finetune training procedure, such as Bert. Pretraining is usually conducted on multiple datasets to learn general abilities like feature extraction. After that, the model is finetuned for a specific task, which may sacrifice its general ability for better performance.

Intuitively, to achieve the best result, a model should be trained in the following way: (i) use MEmeL during pretraining to decrease monosemanticity and increase the general ability, and (ii) exclude MEmeL during finetuning on a specific task to allow for an increase in monosemanticity and achieve better results. To illustrate, pretraining is similar to a student accumulating knowledge from multiple courses, starting from elementary school to university, while finetuning is akin to preparing for the GRE exam. Monosemanticity is similar to rote memorization, where memorizing questions and answers during courses does not help improve the GRE score, but it is useful when studying for the exam.

However, validating the idea and performing pretraining is much more computationally expensive compared to finetuning, especially for large models. For instance, using the naive method to pretrain LLAMA2-7B based on our A800 is estimated to cost over \$$450k$ and over \textbf{5800 GPU-days} \cite{LLAMA2} with our preliminary test. Emergence occurs primarily near $10^{22}$ FLOPs and costs more than \textbf{10500 GPU-days} \cite{DBLP:journals/tmlr/WeiTBRZBYBZMCHVLDF22}. Conducting corresponding validation experiments is impossible for most researchers without the help of top companies.

Under such an arms race on GPUs, our experiments explore our resources to first validate the generality and effectiveness of our methods. Though finetuning may not fully utilize the capabilities of MEmeL, our main experiments demonstrate that finetuning with MEmeL is powerful enough to improve results. Additionally, on the precipitation forecasting task with a smaller scale, we conduct pretraining with MEmeL and observe a significant improvement compared to finetuning. 
The details are provided in subsection.~\ref{ssec:exp:naive}.
This finding supports our idea that compared to finetuning, pretraining can leverage the strengths of MEmeL better. 

\vspace{-4pt}

\subsection{Potential Directions of Emergence Learning}
\citet{DBLP:journals/tmlr/WeiTBRZBYBZMCHVLDF22} raise the concept of emergence in AI models and point out that the explanation for emergence is still under investigation. They discuss the opinion that AI models can be seen as compressors and explain emergence as a result of the accumulation to compress world knowledge. In this aspect, decreasing monosemanticity encourages models to compress knowledge rather than storing knowledge that compromises generality.

\vspace{-4pt}

\subsection{Partially Decreasing Monosemanticity} 
Increasing monosemanticity can be beneficial in some areas where precise memorization is necessary. It is important to study the impact of monosemanticity on different tasks and inhibit or increase it according to the analysis. A complex task could also be broken down into smaller steps, each with its own preference for monosemanticity. 

For example, when asking a model to imitate the author of ``Sapiens: A Brief History of Humankind" and write a preface from scratch, the model should accurately identify Harari as the author through a key-value style memorization and then reproduce the tone in the output with the necessary reasoning ability.


\end{document}